\documentclass[sigconf, nonacm,pdfa]{acmart}

\newcommand\vldbdoi{xx.xxx/xx.xxx}
\newcommand\vldbpages{xxx-xxx}
\newcommand\vldbvolume{xx}
\newcommand\vldbissue{xx}
\newcommand\vldbyear{xx}

\newcommand\vldbauthors{\authors}
\newcommand\vldbtitle{\shorttitle} 
\newcommand\vldbavailabilityurl{https://github.com/real2fish/CSL}
\newcommand\vldbpagestyle{empty}

\usepackage[a-2b]{pdfx}
\usepackage{multirow}
\usepackage{makecell}
\theoremstyle{plain}
\newtheorem{myDef}{Definition}[section]

\usepackage{amsmath}

\usepackage{amssymb}
\usepackage{mathtools}
\usepackage{cleveref}
\usepackage{url}

\usepackage{graphicx}

\usepackage{subcaption}
\usepackage{enumitem}

\usepackage{balance}

\newcommand{\xv}{\boldsymbol{x}}
\newcommand{\Xv}{\boldsymbol{X}}

\newcommand{\zv}{\boldsymbol{z}}
\newcommand{\Zv}{\boldsymbol{Z}}

\newcommand{\eat}[1]{}
\newcommand{\zy}[1]{#1}

\newcommand{\rev}[1]{#1}
\begin{document}
\title{A Shapelet-based Framework for Unsupervised Multivariate Time Series Representation Learning}

\author{Zhiyu Liang}
\affiliation{%
  \institution{Harbin Institute of Technology}
  \city{Harbin}
  \country{China}  
}
\email{zyliang@hit.edu.cn}

\author{Jianfeng Zhang}
\affiliation{%
  \institution{Huawei Noah's Ark Lab}
  \city{Shenzhen}
  \country{China}
}
\email{zhangjianfeng3@huawei.com}

\author{Chen Liang}
\affiliation{%
  \institution{Harbin Institute of Technology}
  \city{Harbin}
  \country{China}
}
\email{1190201818@stu.hit.edu.cn}

\author{Hongzhi Wang}
\affiliation{%
   \institution{Harbin Institute of Technology}
  \city{Harbin}
  \country{China}
  }
\email{wangzh@hit.edu.cn}\authornote{Corresponding author.}

\author{Zheng Liang}
\affiliation{%
 \institution{Harbin Institute of Technology}
  \city{Harbin}
  \country{China}
}
\email{lz20@hit.edu.cn}

\author{Lujia Pan}
\affiliation{%
  \institution{Huawei Noah's Ark Lab}
  \city{Shenzhen}
  \country{China}
}
\email{panlujia@huawei.com}


\begin{abstract}
Recent studies have shown great promise in unsupervised representation learning (URL) for multivariate time series,  because URL has the capability in learning generalizable representation for many downstream tasks without using inaccessible labels.  However, existing approaches usually adopt the models originally designed for other domains (e.g., computer vision) to encode the time series data and {rely on strong assumptions to design learning objectives, which limits their ability to perform well}. To deal with these problems, we propose a novel URL framework for multivariate time series by learning time-series-specific shapelet-based representation through a popular contrasting learning paradigm. To the best of our knowledge, this is the first work that explores the shapelet-based embedding in the unsupervised general-purpose representation learning. A unified shapelet-based encoder and a novel learning objective with multi-grained contrasting and multi-scale alignment are particularly designed to achieve our goal, and a data augmentation library is employed to improve the generalization. We conduct extensive experiments using tens of real-world datasets to assess the representation quality on many downstream tasks, including classification, clustering, and anomaly detection. The results demonstrate the superiority of our method against not only URL competitors, but also techniques specially designed for downstream tasks. Our code has been made publicly available at https://github.com/real2fish/CSL.  
\end{abstract}

\maketitle

\pagestyle{\vldbpagestyle}
\begingroup\small\noindent\raggedright\textbf{PVLDB Reference Format:}\\
\vldbauthors. \vldbtitle. PVLDB, \vldbvolume(\vldbissue): \vldbpages, \vldbyear.\\
\href{https://doi.org/\vldbdoi}{doi:\vldbdoi}
\endgroup
\begingroup
\renewcommand\thefootnote{}\footnote{\noindent
This work is licensed under the Creative Commons BY-NC-ND 4.0 International License. Visit \url{https://creativecommons.org/licenses/by-nc-nd/4.0/} to view a copy of this license. For any use beyond those covered by this license, obtain permission by emailing \href{mailto:info@vldb.org}{info@vldb.org}. Copyright is held by the owner/author(s). Publication rights licensed to the VLDB Endowment. \\
\raggedright Proceedings of the VLDB Endowment, Vol. \vldbvolume, No. \vldbissue\ %
ISSN 2150-8097. \\
\href{https://doi.org/\vldbdoi}{doi:\vldbdoi} \\
}\addtocounter{footnote}{-1}\endgroup

\ifdefempty{\vldbavailabilityurl}{}{
\vspace{.3cm}
\begingroup\small\noindent\raggedright\textbf{PVLDB Artifact Availability:}\\
The source code, data, and/or other artifacts have been made available at \url{\vldbavailabilityurl}.
\endgroup
}

\section{Introduction}\label{sec:intro}
Multivariate time series (MTS) generally describes a group of dependent variables evolving over time, each of which represents a monitoring metric (e.g., temperature or CPU utilization) of an entity (e.g., system or service). MTS data play a vital role in many practical scenarios, such as manufacturing, medicine, and finance~\cite{kim2019fault,riley2021internet,bennett2022detection}\eat{\footnote{references}}. 

While MTS are being increasingly collected from various applications, a particular challenge in modeling them is the lack of labels. Unlike images or text that usually contain human-recognizable patterns, label acquisition for time series is much more difficult, because the underlying state of these time-evolving signals can be too complicated even for the domain specialists~\cite{TNC}. For this reason, it has recently become a research focus to explore unsupervised (a.k.a. self-supervised) representation learning (URL)\eat{\footnote{any references?}} for MTS~\cite{TST,TS-TCC,ts2vec,BTSF}. {\textit{URL aims to train a neural network (called encoder) without accessing the labels to embed the data into feature vectors, by using carefully designed learning objectives to leverage the inherent structure of the raw data.} The learned representations (a.k.a. features or embeddings) can then be used for training models to solve a downstream analysis task using \textit{little annotated data} compared to the traditional supervised methods~\cite{TST}. And the features are more \textit{general-purpose} since they can facilitate to several tasks.}



Unfortunately, unlike in domains such as computer vision (CV)\cite{SimCLR,PCL,CV_repr_Univip} or natural language processing (NLP)~\cite{SimCSE,nlp_repr_zhang2022unsupervised}, URL in the context of time series is still under-explored. MTS are typically continuous-valued data with \rev{high noise, diverse temporal patterns and varying semantic meanings}, etc~\cite{UEA}. These unique complexities make advanced URL methods in the aforementioned domains difficult to perform well ~\cite{TNC, TS-TCC}. Although several studies have attempted to fill this gap by considering the characteristics of time series, such as the \rev{time-evolving nature~\cite{TNC}} and the multi-scale semantics~\cite{ts2vec, T-Loss}, \textit{existing approaches can still be weak in learning well-performed representations} partly due to the following reasons. 

First,\eat{the encoder (i.e., the parameterized model that maps raw data to representations) design of existing frameworks} the existing representation encoder designs are highly inspired by experiences in CV and NLP domains, which may not be well-suited for MTS. Specifically, {convolutional neural network (CNN)~\cite{causal_conv_wavenet,resnet} and Transformer~\cite{attention} are commonly-used encoders in recent studies~\cite{ts2vec,TS-TCC,TST,TNC,T-Loss,BTSF}. However, the encoders still face many difficulties when applying in MTS due to the lack of capability to deal with the characteristics of time series ~\cite{OSCNN,FEDformer,Pyraformer}.} Second, some existing approaches rely on domain-specific assumptions, such as the neighbor similarity~\cite{T-Loss,TNC} and the contextual consistency~\cite{ts2vec}, thus are difficult to generalize to various scenarios. For instance, ~\citet{T-Loss} and \citet{TNC} assume that subsequences distant in time should be dissimilar, which can be easily violated in periodic time series~\cite{sanei2013eeg}.


To tackle the issues mentioned above, we explore the time-series-specific representation encoder without strong assumptions for URL. In particular, we consider the encoder based on a non-parametric time series analysis concept named shapelet~\cite{ye2011time}, i.e. salient subsequence which is tailored to \rev{extract time series features from only important time windows to avoid the noises outside}. \zy{The main reason is that the shapelet-based representation has shown superior performance in specific tasks such as classification~\cite{ma2019triple,icde22shapelets,Shapenet} and clustering~\cite{MUSLA}. Besides, compared to the feature extracted from other neural networks such as CNN, the shapelet-based feature can be more intuitive to understand~\cite{ye2011time}.} However, it has never been explored in the recently rising topic of URL for general-purpose representation.  To fill this gap, we take the first step and propose to \textit{learn shapelet-based encoder employing contrastive learning}, a popular paradigm that has shown success in URL~\cite{SimCLR,TS-TCC,ts2vec,TF-C}.

\textit{We highlight three challenges in learning high-quality and general-purpose shapelet-based representation.} The first is how to design a shapelet-based encoder to capture diverse temporal patterns of various time ranges, considering that it is originally proposed to represent only a single shape feature, and \rev{exhaustive search} or prior knowledge is needed to determine the encoding scale~\cite{bostrom2017binary,Shapenet,MUSLA}. The second is how to design a URL objective to learn \rev{general information for downstream tasks} through this shapelet-based encoder, which has never been studied. Last, while contrastive learning leverages the \rev{representation similarity} of the augmentations of one sample~\cite{SimCLR} to learn the encoder, it remains an open problem to properly augment the time series \rev{to keep the similarity}~\cite{ts2vec,TNC}.

To cope with these challenges, we propose a novel unsupervised MTS representation learning framework named \textit{\textbf{C}ontrastive \textbf{S}hapelet \textbf{L}earning (CSL)}. Specifically, we design a unified architecture that uses multiple shapelets with various (dis)similarity measures and lengths to jointly encode a sample, \rev{such that to capture diverse temporal patterns from short to long term}. As shapelets of different lengths can separately embed one sample into different representation spaces that are complementary with each other, we propose a multi-grained contrasting objective to simultaneously consider {the joint embedding and the representations at each time scale. In parallel, we design a multi-scale alignment loss} to encourage the representations of different scales to achieve consensus. \rev{The basic idea is to automatically capture the varying semantics by leveraging the intra-scale and inter-scale dependencies of the shapelet-based embedding.} Besides, we develop an augmentation library using diverse types of data augmentation methods to further improve the representation quality. To the best of our knowledge, CSL is the first general-purpose URL framework based on shapelets. The main contributions are summarized as follows:

\begin{itemize}[leftmargin=5ex,nosep,topsep=4pt]
\item This paper studies how to improve the URL performance using time-series-specific shapelet-based representation, which has achieved success in specific tasks but has never been explored for the general-purpose URL.
    \item A novel framework is proposed that adopts contrastive learning to learn shapelet-based representations. A unified shapelet-based encoder architecture and a learning objective with multi-grained contrasting and multi-scale alignment are particularly designed to capture diverse patterns in various time ranges. A library containing various types of data augmentation methods is constructed to improve the representation quality.   
    \item Experiments on tens of real-world datasets from various domains show that i) our learned representations are general to many downstream tasks, such as classification, clustering, and anomaly detection; ii) the proposed method outperforms existing URL competitors and can be comparable to (even better than) tailored techniques for classification and clustering. Additionally, we study the effectiveness of the key components proposed in CSL and the model sensitivity to the key parameters, demonstrate the superiority of  CSL against the fully-supervised competitors on partially labeled data, and explain the shapelets learned by CSL. \rev{We also study our method in long time series representation and assess its running time.}       
\end{itemize}

\section{Related Work}\label{sec:related_work}
There are two lines of research  closely related to this paper: 

\eat{This section briefly reviews the related work on unsupervised MTS representation learning and time series shapelets.}

\noindent
\textbf{Unsupervised MTS representation learning.} 
Unlike in domains such as CV~\cite{SimCLR,PCL,CV_repr_Univip,yang2022vision} and NLP~\cite{SimCSE,nlp_repr_zhang2022unsupervised}, the study of URL in time series is still in its infancy. 

Inspired by word representation~\cite{word2vec}, \citet{T-Loss} adapts the triplet loss to time series to achieve URL. Similarly, \citet{TST} explores the utility of transformer~\cite{attention} for URL due to the success of transformer in modeling natural language. \citet{CPC} proposes to learn the representation by predicting the future in latent space. \citet{TS-TCC} extends this idea by conducting both temporal and contextual contrasting to improve the representation quality. Instead of using prediction, ~\citet{ts2vec} combines timestamp-level contrasting with contextual contrasting to achieve hierarchical representation. \citet{TNC} assumes consistency between overlapping temporal neighborhoods to model dynamic latent states, while \citet{BTSF} utilizes the consistency between temporal and spectral domains to enrich the representation. Although these methods have achieved improvements in representation quality, they still have limitations such as the lack of intuitions in encoder design and the dependency on specific assumptions, as discussed in Section~\ref{sec:intro}.

\noindent
\textbf{Time series shapelet.} The concept of shapelet is first proposed by \citet{ye2011time} for supervised time series classification tasks. It focuses on extracting features in a notable time range to reduce the interference of noise, which is prevalent in time series. 

In the early studies, shapelets are selected by enumerating subsequences of the training time series~\cite{ye2011time, mueen2011logical,hills2014classification,bostrom2017binary}, which suffers from non-optimal representation and high computational overhead~\cite{grabocka2014learning}. To address these problems, a shapelet learning method is first proposed by \citet{grabocka2014learning}, which directly learns the optimal shapelets through a supervised objective. After this study, many approaches~\cite{ma2019triple,ma2020adversarial, liang2021efficient,icde22shapelets} have been proposed to improve the effectiveness and efficiency for classification. Except for supervised classification task, some works~\cite{u-shapelets,USSL,MUSLA} employ shapelets for time series clustering and also show competitive performance. 

ShapeNet~\cite{Shapenet} is a special work related to both URL and shapelet. However, \rev{it aims to ``select'' shapelets from existing candidates for MTS classification, while it just adopts a CNN-based URL method extended from~\cite{T-Loss} to assist the selection. It even contains a supervised feature selection step which uses the true labels. Instead, both our CSL and other URL methods target a different problem that is to automatically ``learn'' the new features not present in existing feature set without using labels to tackle more than one task.}

In summary, although shapelet-based representation has been widely studied for classification and clustering tasks, it has never been explored for the unsupervised learning of general-purpose representations facilitating various tasks as our CSL.

\section{Problem Statement}
This section defines the key concept used in the paper. At first, we define the data type we are interested in, \textit{multivariate time series}.

\vspace{-1ex}
\begin{myDef}[Multivariate Time Series]
{Multivariate time series (MTS) is a set of variables, each including observations ordered by successive time.} Formally, we denote a multivariate time series sample with $D$ variables (a.k.a. dimensions or channels) and $T$ timestamps (a.k.a. length) as $\boldsymbol{x} \in \mathbb{R}^{D \times T}$, and a dataset containing $N$ samples as $\boldsymbol{X} = \{\boldsymbol{x}_1, \boldsymbol{x}_2, ..., \boldsymbol{x}_N\} \in \mathbb{R}^{N \times D \times T}$. 
\end{myDef}
\vspace{-1ex}

Then, the problem that we are addressing, i.e., \textit{unsupervised representation learning for MTS}, is formulated as follows.
\vspace{-1ex}


{
\begin{myDef}[Unsupervised Representation Learning for MTS]
Given an MTS dataset $\boldsymbol{X}$, the goal of unsupervised representation learning (URL) is to train a neural network model (encoder) $f: \mathbb{R}^{D \times T} \mapsto \mathbb{R}^{D_{repr}}$, such that the representation $\boldsymbol{z}_i = f(\xv_i)$ can be informative for downstream tasks, e.g., classification and anomaly detection\eat{, and yield performance comparable to its supervised counterparts\eat{that of parallel supervised paradigms}}. Here \textit{unsupervised} means that the labels of downstream tasks are unavailable when training $f$. To simplify the notation, we denote  $\boldsymbol{Z} = f(\boldsymbol{X}) = \{\boldsymbol{z}_1, \boldsymbol{z}_2, ..., \boldsymbol{z}_N\}$ in following sections.
\end{myDef}
}

\vspace{-1ex}
\zy{It is worthy to note that some works limit ``unsupervised (representation) learning'' to the unsupervised tasks (e.g., clustering~\cite{MUSLA}), so the competitors are only the unsupervised methods. Instead, the URL problem mentioned in this paper is to \textit{learn the features that can not only tackle the unsupervised tasks, but also achieve comparable performance to the supervised competitors on the classification task,} which can be more general yet challenging. }

\section{Methodology}
In this section, the proposed framework and all components \rev{are elaborated}.

\vspace{-1ex}
\subsection{Overview}
\begin{figure}[t]
    \centering
    \includegraphics[width=\linewidth]{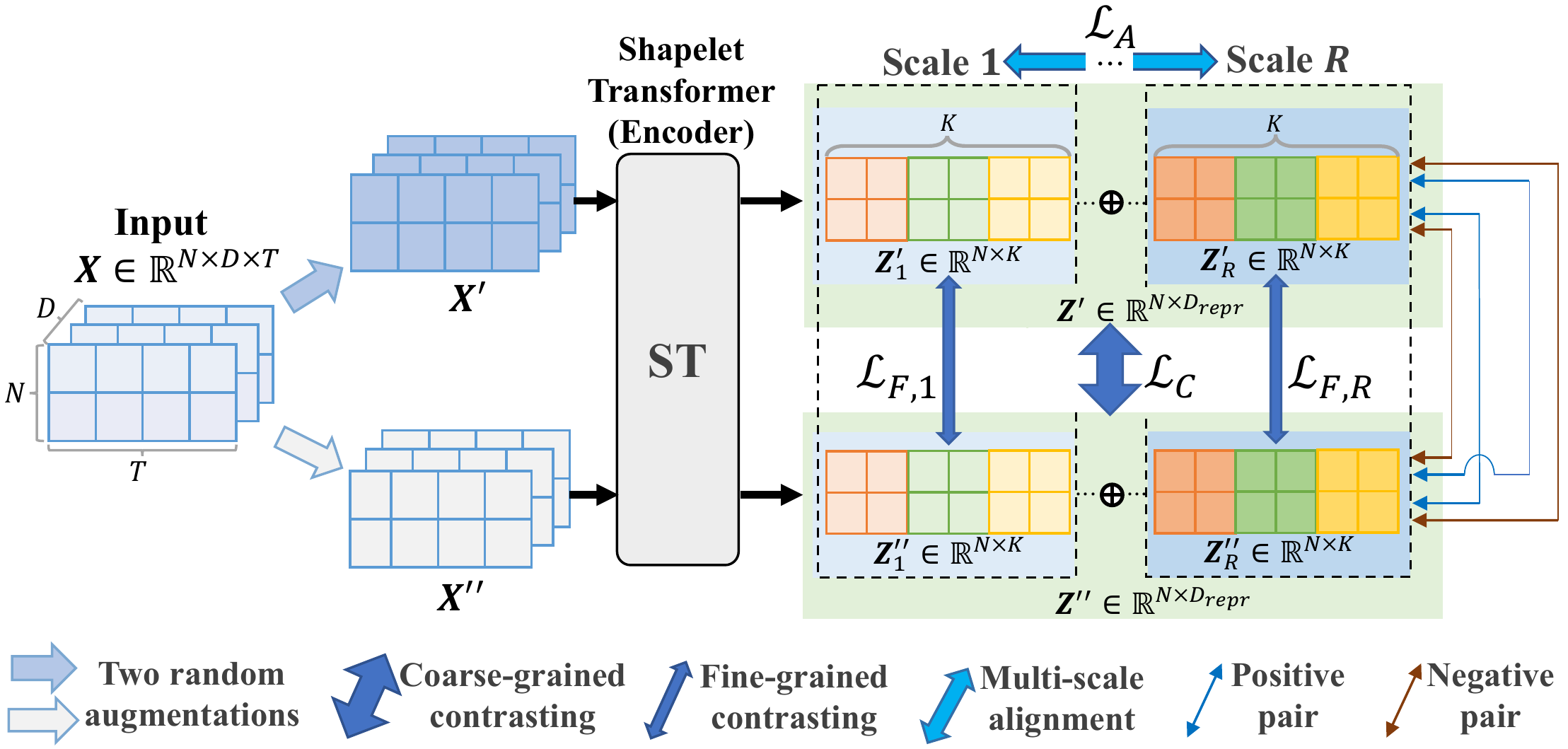}
    \vspace{-4ex}
    \caption{Overview framework of CSL.}
    \label{fig:framework}
    \vspace{-3ex}
\end{figure}

{We illustrate the overview framework of the proposed contrastive shapelet learning (CSL) in Fig.~\ref{fig:framework}.} 
Given the input $\boldsymbol{X}$, \rev{two data augmentation methods, denoted as $A^\prime(\xv)$ and $A^{\prime\prime}(\xv)$, are randomly selected from a library (to be discussed later) to produce two correlated views of $\Xv$ as $\boldsymbol{X}^{\prime} = A^\prime(\Xv)$ and $\boldsymbol{X}^{\prime\prime}=A^{\prime\prime}(\Xv)$, where $A^\prime(\Xv) = \{A^\prime(\xv_1), \ldots, A^\prime(\xv_N)\}$ and the same to $A^{\prime\prime}(\Xv)$.} Then these two views are fed into a time-series-specific encoder named Shapelet Transformer (ST), which embeds the samples into a latent space (see Section~\ref{sec:shapelet_transformer}). CSL explores the representation in this latent space where different shapelets serve as the basis (see Section~\ref{sec:multi-grained-contrasting} and \ref{sec:multi-scale alignment}). We believe that our method is more general as it does not depend on task-specific assumptions like ~\cite{T-Loss,TNC,ts2vec}.

Formally, given $\boldsymbol{X}^{\prime}$ and $\boldsymbol{X}^{\prime\prime}$, we have the shapelet-based representations as:
\vspace{-1ex}
\begin{equation}
\begin{split}
    \boldsymbol{Z}^{\prime} &= f(\boldsymbol{X}^{\prime}) \in \mathbb{R}^{N \times D_{repr}}, \\
    \boldsymbol{Z}^{\prime\prime} &= f(\boldsymbol{X}^{\prime\prime}) \in \mathbb{R}^{N \times D_{repr}}.
\end{split}
\end{equation}
\vspace{-2ex}

Following the paradigm of contrastive learning~\cite{SimCLR,TF-C,TS-TCC}, for each $\boldsymbol{x}_{i}$, the embedding $\boldsymbol{z}^{\prime}_{i}$ should be close to $\boldsymbol{z}^{\prime\prime}_{i}$ whereas far away from  $\boldsymbol{z}^{\prime\prime}_{j}$ derived from other samples where $j \neq i$. The encoder is learned through maximizing the similarity of the positive pairs $(\boldsymbol{z}^{\prime}_{i}, \boldsymbol{z}^{\prime\prime}_{i})$ and minimizing the similarity of the negative pairs $(\boldsymbol{z}^{\prime}_{i}, \boldsymbol{z}^{\prime\prime}_{j})$. \rev{Note that using data augmentation to generate the positive pairs is a common way for contrastive learning which is required by most URL methods, including TS2Vec~\cite{ts2vec}, TS-TCC~\cite{TS-TCC}, etc~\cite{BTSF,TST}.  Alternatively, T-Loss~\cite{T-Loss} and TNC~\cite{TNC} select subsequences as positive samples. Both augmented and sampled time series serve as the self-supervised signals in URL, which plays the similar role as the labels used by the supervised methods (e.g., OSCNN~\cite{OSCNN}).} 

\begin{figure}[t]
    \centering
\subcaptionbox{\centering Origin.
    \label{subfig:original}}
{\includegraphics[width=.3\linewidth]{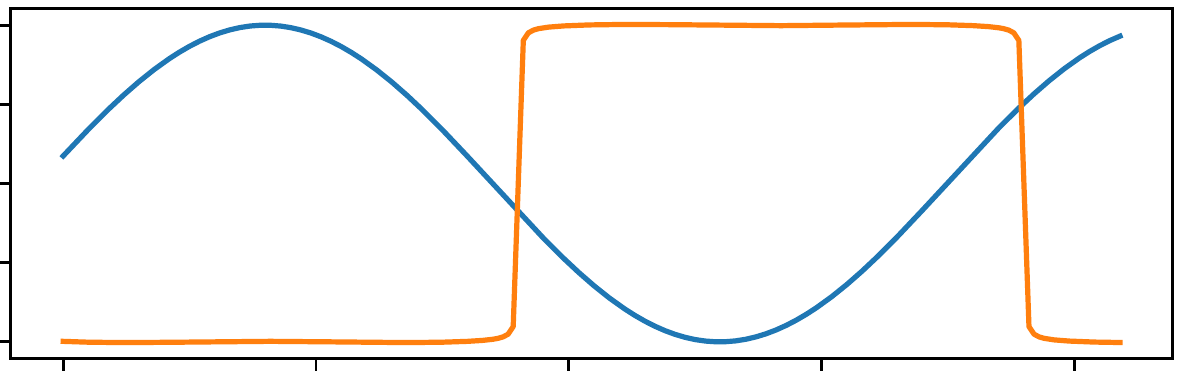}\vspace{-1ex}}
\subcaptionbox{\centering Jittering.
    \label{subfig:jitter}}
{\includegraphics[width=.3\linewidth]{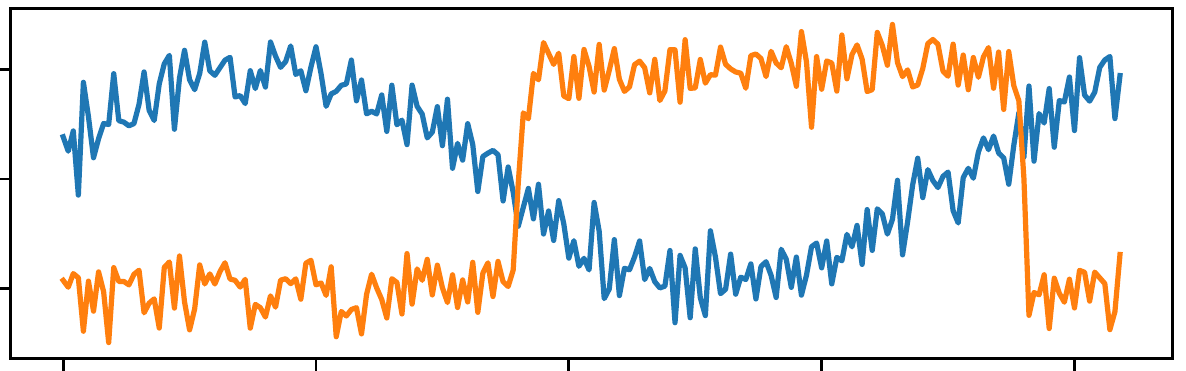}\vspace{-1ex}}
\subcaptionbox{\centering Cropping.
    \label{subfig:crop}}
{\includegraphics[width=.3\linewidth]{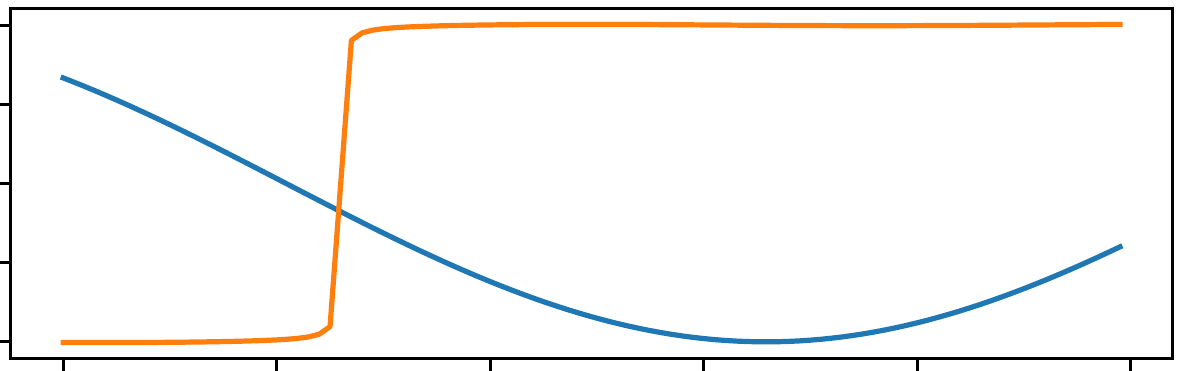}\vspace{-1ex}}
\subcaptionbox{\centering Time warping.
    \label{subfig:tw}}
{\includegraphics[width=.3\linewidth]{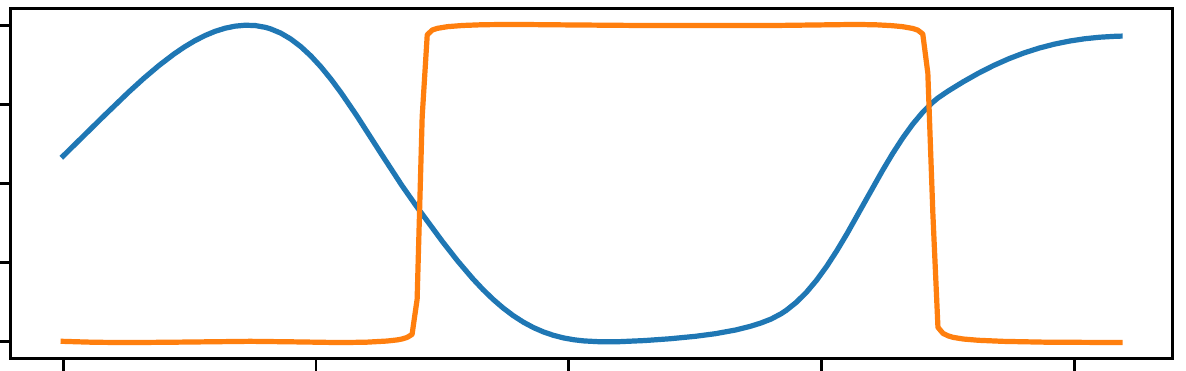}\vspace{-1ex}}
\subcaptionbox{\centering Quantizing.
    \label{subfig:quantize}}
{\includegraphics[width=.3\linewidth]{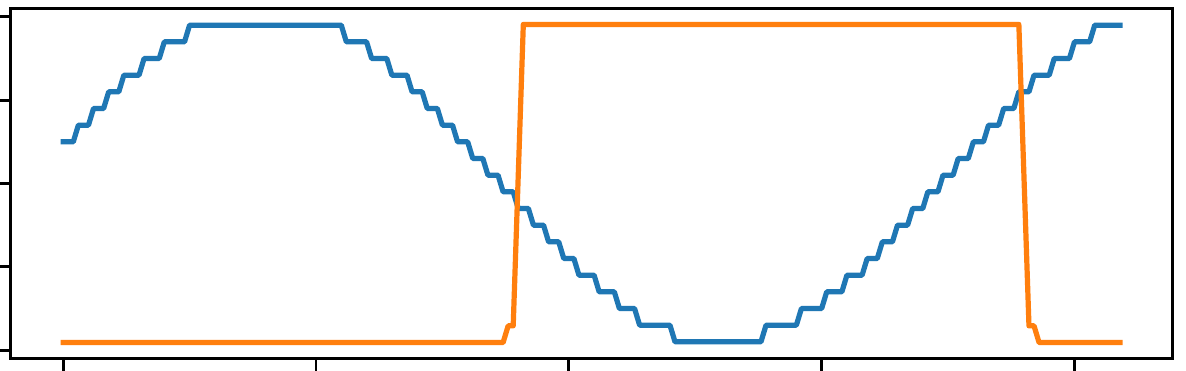}\vspace{-1ex}}
\subcaptionbox{\centering Pooling.
    \label{subfig:pool}}
{\includegraphics[width=.3\linewidth]{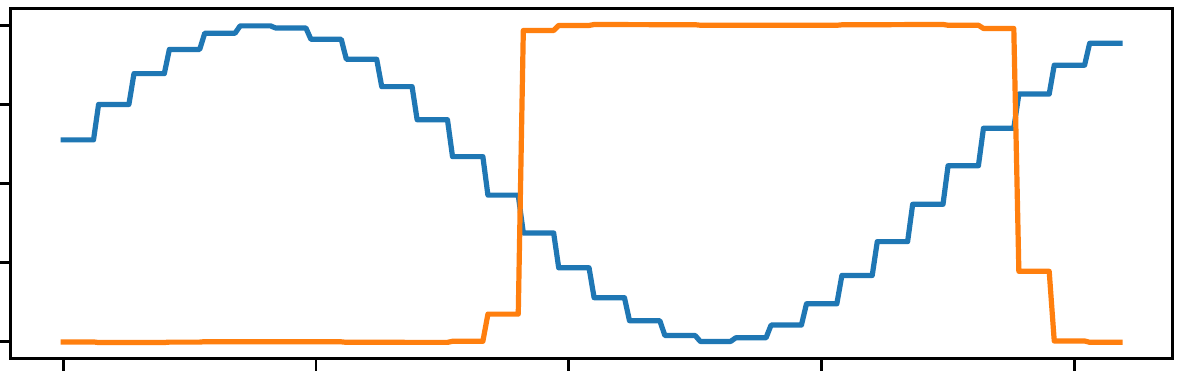}\vspace{-1ex}}
\vspace{-2ex}
\caption{Illustration of the data augmentation methods using a two-dimensional time series. All methods are identically performed on each dimension of the original time series.}
\label{fig:illustration_aug}
\vspace{-3ex}
\end{figure}

Despite the success of contrastive learning in URL~\cite{TS-TCC,ts2vec,BTSF,TF-C}, \rev{an open question is how to determine proper data augmentation methods to ensure representation similarity of positive samples~\cite{SimCLR}, which could be data- and model-dependent~\cite{survey_aug_oneplus}.} It is beyond the scope of this paper to develop new augmentation techniques or augmentation selection algorithms. \zy{Instead, we construct a data augmentation library which contains \textit{diverse types of methods} for the random selection at each training step (illustrated in Fig.~\ref{fig:framework}), so that they can be complementary with each other to adapt to various time series data. The library consists of five well-established time series augmentation methods, including  \textit{jittering} $J(\xv)$ that adds random noise to each observation, \textit{cropping} $C(\xv)$ that crops the time series into a randomly selected subsequence, \textit{time warping} $TW(\xv)$ that stretches or contracts the randomly selected subsequences, \textit{quantizing} $Q(\xv)$ that quantizes each observation to the nearest level, and \textit{pooling} $P(\xv)$ that reduces the temporal resolution using average pooling on each consecutive observations. We illustrate how they are performed using the running examples in Fig.~\ref{fig:illustration_aug}, and we refer interested readers to~\cite{data_aug_wearable,tsaug} for more details.}  

{The encoder ST is designed to capture the patterns of different time scales using separated shapelets with different lengths. Thus, we propose a multi-grained contrasting objective to simultaneously perform contrastive learning on the shapelet-based embedding of every single scale (fined-grained contrasting) and the representations in the joint space $\mathbb{R}^{D_{repr}}$ regarding all scales (coarse-grained contrasting). Additionally, \zy{inspired by the consensus principle in multi-view learning~\cite{multiview-survey}, we design a multi-scale alignment term to encourage the features at different scales to achieve agreement.}}


In the rest of this section, \rev{we elaborate} the key components of the proposed CSL framework, {including \textit{Shapelet Transformer}, \textit{multi-grained contrasting}, and \textit{multi-scale alignment}.}\eat{\footnote{connecting to the following subsections?}}

\subsection{Shapelet Transformer}\label{sec:shapelet_transformer}

Shapelet is originally designed to extract representative shape features of univariate time series~\cite{ye2011time}. \zy{In this paper, we simply extend it to a more general multivariate case.} Given a sample $\boldsymbol{x} \in \mathbb{R}^{D \times T}$, a multivariate shapelet $\boldsymbol{s} \in \mathbb{R}^{D \times L} (L < T)$ \zy{which has the same dimension $D$} encodes $x$ using the Euclidean norm between $\boldsymbol{s}$ and the best-matching subsequence relative to $\boldsymbol{s}$ within $\boldsymbol{x}$, defined as: 
\begin{equation}
dist(\boldsymbol{x}, \boldsymbol{s}) = \mathop{\min}_{t = 1,2,...,T-L+1} ||\boldsymbol{x}[t,L] -\boldsymbol{s}||_2  \in \mathbb{R},
\label{eq:shapelet_distance}
\end{equation}
where $\boldsymbol{x}{[t,L]}$ denotes the subsequence of $\boldsymbol{x}$ starting at timestamp $t$ and lasting $L$ steps. \zy{By taking $\boldsymbol{s}$ as \textit{trainable parameter}, we can directly learn the optimal shapelet like \rev{any neural network} using the optimization algorithm such as stochastic gradient descent (SGD)~\cite{ma2019triple}.}   However, \textit{it is difficult to capture patterns beyond shapes for the original definition in Eq.~\eqref{eq:shapelet_distance},} such as the spectral information in the frequency domain, which can limit the capability of the shapelet-based encoder. To address this problem, we extend the representation to a general form as:
\begin{equation}
g(\boldsymbol{x}, \boldsymbol{s}, d) = \mathop{\mathrm{agg}_{d}}_{t = 1,2,...,T-L+1} \sum_{j=1}^D d(\boldsymbol{x}^{j}[t,L], \boldsymbol{s}^j) \in \mathbb{R},
\label{eq:general_shapelet}
\end{equation}
where $\boldsymbol{x}^j$ ($\boldsymbol{s}^j$) represents the series (shapelet) at $j$-th dimension, and $\mathrm{agg}_{d}$ is the aggregator that produces the result of $d$ between the most similar  pair of ($\boldsymbol{x}[i,L]$, $\boldsymbol{s}$). $d(\cdot, \cdot)$ can be \textit{any (dis)similarity measure} for equal-length series.  Eq.~\eqref{eq:shapelet_distance} is obviously a special case of Eq.~\eqref{eq:general_shapelet} when $d$ is Euclidean norm, and $g$ corresponds to 1-D convolution when $d$ is the cross-correlation function~\cite{cross-correlation}.

\begin{figure}[t]
    \centering
    \includegraphics[width=0.7\linewidth]{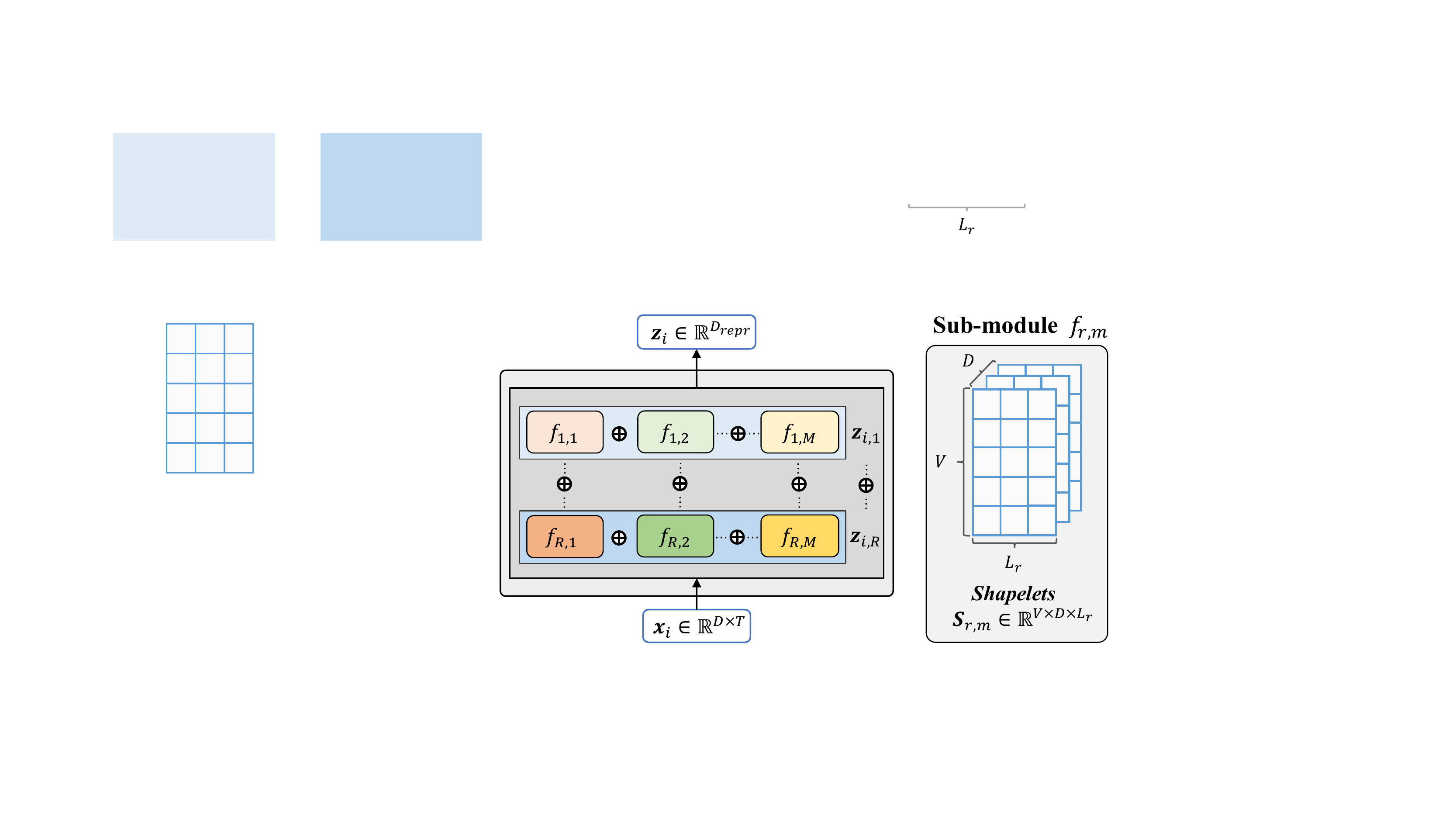}
    \vspace{-2ex}
    \caption{Architecture of Shapelet Transformer (ST)}
    \label{fig:encoder}
    \vspace{-2ex}
\end{figure}

Based on \eat{the general formulation}Eq.~\eqref{eq:general_shapelet}, we design a shapelet-based encoder named \textit{Shapelet Transformer} (ST), which is shown in Fig.~\ref{fig:encoder}.\eat{Note that our Shapelet Transformer is independent of the Transformer neural network~\cite{attention}. We name the encoder as Shapelet Transformer because it ``transforms'' the time series into representations based on the shapelets and the process has been called ``shapelet transformation'' in the literature~\cite{shapelet-transformation}. We clarify this issue to avoid ambiguity.} To extract diverse temporal patterns, ST is a combination of multiple sub-modules with shapelets of $R$ various lengths (scales) and $M$ different (dis)similarity measures. The core idea comes from the {observations}\eat{experience} that i) time series could possess both short-term and long-term patterns in practice~\cite{OSCNN,EGS}, and ii) different measures can be complementary with each other to produce more informative features~\cite{hive-cote}. However, our design is eventually different from existing approaches since we \textit{simultaneously consider these two aspects in a unified shapelet-based architecture.} 

We denote the sub-module with shapelets of length $L_r$ (scale $r$) and (dis)similarity measure $d_m$ as $f_{r, m}$. Each $f_{r,m}$ has $V$ shapelets $\boldsymbol{s}_{r,m,1}, \ldots, \boldsymbol{s}_{r,m,V}$ that separately embed the input. The outputs of all sub-modules for one sample are concatenated to jointly represent the sample. Formally, the encoder ST is defined as: 
\begin{equation}
\begin{split}
\zv_i =&
f(\boldsymbol{x}_i) = \boldsymbol{z}_{i,1} \oplus \boldsymbol{z}_{i,2} \oplus \ldots \oplus \boldsymbol{z}_{i,R}, \\
\boldsymbol{z}_{i,r} =& f_{r,1}(\xv_i) \oplus  f_{r,2}(\xv_i) \oplus \ldots \oplus f_{r,M}(\xv_i),\\
    f_{r,m}(\xv_i) =& [f_{r,m,1}(\xv_i), f_{r,m,2}(\xv_i), \ldots , f_{r,m,V}(\xv_i)],
\end{split}
\end{equation}
where {$\oplus$ is the concatenation operator}, $\boldsymbol{z}_{i,r} \in \mathbb{R}^{K}(K=MV)$ denotes the representation of $\boldsymbol{x}_i$ at scale $r$, and $ f_{r,m,v}(\boldsymbol{x}_i) = g(\boldsymbol{x}_i,\boldsymbol{s}_{r,m,v}, d_m)$. 

Note that although extracting multi-scale features is a widely adopted idea for time series~\cite{OSCNN,EGS}, \textit{the proposed Shapelet Transformer provides a simple yet effective way to achieve contrastive learning for the features of different scales (Section~\ref{sec:multi-grained-contrasting}) and to maximize the agreement of the scales (Section~\ref{sec:multi-scale alignment})}, which we show essential for improving the performance (Section~\ref{sec:exp:ablation}). The reason is that we simply concatenate the features encoded by each of the shapelets, and thus the representations of different scales can be separated from each other, while the features of existing multi-scale networks~\cite{OSCNN,EGS} are usually fused through complicated layer-by-layer structures. Moreover, integrating multiple measures into the multi-scale shapelet-based architecture is also a simple yet effective idea for improving the URL performance (Section~\ref{sec:exp:ablation}). 

Shapelet Transformer is a unified architecture that can be flexibly changed by varying $R$, $M$, $V$, $L_r$, $d_m$ if one has prior knowledge,  such as the time scale of the MTS patterns. Considering that prior knowledge is not always easy to access, we introduce \textit{a general configuration of the model structure.} Specifically, we fix $R$ to a moderate value of 8 and adaptively set $L_r$ as \eat{$L_r = r\times 0.1T$} the evenly spaced numbers over $[0.1T, 0.8T]$, i.e., $L_r = r \times 0.1T$ ( $r \in \{1,\ldots,R\}$), to approximately match the patterns from short to long term. 
{Three widely adopted (dis)similarity measures are considered, including \textit{Euclidean distance} ($d_1$), \textit{cosine similarity} ($d_2$) and \textit{cross correlation} ($d_3$).} 
As such, given the dimension $D_{repr}$ of the output embedding, the encoder structure can be automatically determined. 


\subsection{Multi-grained Contrasting}\label{sec:multi-grained-contrasting}
After encoding the training samples, we employ contrastive learning, a popular paradigm that has achieved success in URL,  to learn the Shapelet Transformer. The principle of contrastive learning is to pull close the positive pairs and push apart the negative pairs in the embedding space. In this paper, we adopt the InfoNCE loss~\cite{CPC} to separate positive from negative samples \zy{because it is one of the most popular loss functions in contrastive learning which has been widely shown effective~\cite{CPC,Moco,gcc}, but the contrastive loss in any other form can also fit in our framework.}  Given an embedding $\zv_i$ of a sample $\boldsymbol{x}_i$ and a set $\boldsymbol{Z}$ that contains the embeddings of one positive sample $\boldsymbol{x}_i^+$ and $N - 1$ negative samples of $\boldsymbol{x}_i$, the contrastive loss is defined as: 
\begin{equation}
    \mathcal{L}_{IN}(z_i, \boldsymbol{Z}) = - \log \frac{\exp(\mathrm{sim}(\zv_i, \zv_i^+)/\tau)}{\sum_{\boldsymbol{z}^\prime \in \boldsymbol{Z}} \exp(\mathrm{sim}(\zv_i, \boldsymbol{z}^\prime)/\tau)}\label{eq:infonce},
\end{equation}
where $\zv_i^+$ is the embedding of $\boldsymbol{x}_i^+$, $\mathrm{sim}(\boldsymbol{u}, \boldsymbol{v}) = \boldsymbol{u} \cdot \boldsymbol{v} / \|\boldsymbol{u}\| \|\boldsymbol{v}\|$ is the cosine similarity, and $\tau$ is a temperature parameter that controls the strength of penalties on hard negative samples~\cite{CPC}. 

\begin{figure}[t]
    \centering
    \includegraphics[width=.9\linewidth]{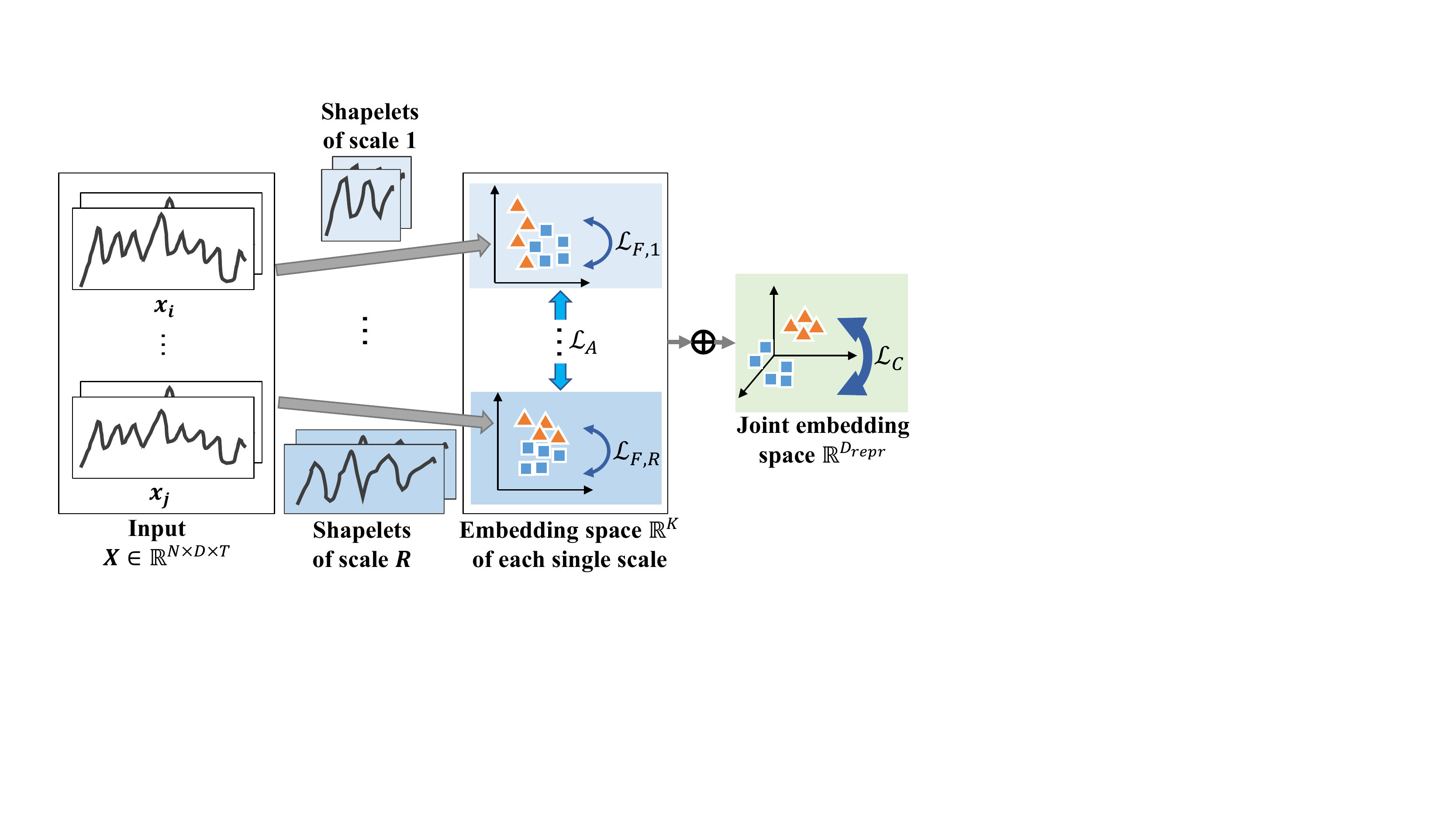}
    \vspace{-2ex}
    \caption{Illustration of multi-grained contrasting and multi-scale alignment. Display one shapelet at each scale for clarity.}
    \label{fig:learning_objective}
    \vspace{-2ex}
\end{figure}

Recall that our Shapelet Transformer embeds one MTS sample into $R$ separate embedding spaces which capture temporal features at different time scales. Thus, we propose a multi-grained contrasting objective that \textit{explicitly considers not only the joint embedding space of the $R$ scales, but also the latent space for each single scale}, as illustrated in Fig.~\ref{fig:learning_objective}. Specifically, we consider contrastive learning in the joint embedding space of all scales as the coarse-grained contrasting. The loss is defined as:
\begin{equation}
    \mathcal{L}_C = \sum_{i=1}^N \mathcal{L}_{IN}(\boldsymbol{z}^{\prime}_{ i}, \boldsymbol{Z}^{\prime\prime}).
\end{equation}
In parallel, the fine-grained contrasting is performed for embedding at each time scale $r \in \{1,...,R\}$, defined as:
\begin{equation}
    \mathcal{L}_{F,r} = \sum_{i=1}^N \mathcal{L}_{IN}(\boldsymbol{z}^{\prime}_{ i, r}, \boldsymbol{Z}^{\prime\prime}_{r}).
\end{equation}
The multi-grained contrastive loss is the sum of the coarse-fined and the fine-grained loss functions, i.e.,
\begin{equation}
    \mathcal{L}_M = \mathcal{L}_C + \sum_{r=1}^R \mathcal{L}_{F,r}.
\end{equation}

\zy{One may wonder \textit{why the coarse-grained contrasting is required given that the optimal representations at each single scale seem to be learned using the fine-grained losses.} The intuition is that with only the fine-grained contrasting, the learning process could be dominated by some scales, i.e., the embeddings of some scales are well learned but the others are not,   such that the joint embedding is not optimal. Thus, we design the coarse-grained loss to explicitly encourage feature similarity in the joint embedding space, so that to ``balance'' the representation quality of each scale to improve the final performance. We further clarify this issue in Section~\ref{sec:exp:ablation}.}

\subsection{Multi-scale Alignment}\label{sec:multi-scale alignment}
As illustrated in Fig.~\ref{fig:learning_objective}, the joint embedding space $\mathbb{R}^{D_{repr}} (D_{repr} = RK)$ is composed by the space $\mathbb{R}^{K}$ of each single scale. \zy{For one time series, the features of different scales are extracted using shapelets of different lengths, and thus can be seen as different views of the sample (similar to images of a 3-D object taken from different viewpoints~\cite{multiview-survey}). From the perspective of multi-view learning, the representations of different lengths have not only complementary information but also consensus because the shapelets of different lengths can match some correlated time regions.} 
\zy{Since we have leveraged the complementary information by using the joint embeddings, we propose to \textit{enhance the consensus over different scales}, which can help to reduce the error rate of each view (scale)~\cite{multiview-survey}. }

Inspired by Canonical Correlation Analysis (CCA)~\cite{deep_CCA}, we design a multi-scale alignment strategy to promote the consensus. The basic idea is to \textit{encourage the embeddings of one sample for different scales to be maximally correlated.} Formally, given representations $\boldsymbol{Z}_1, \boldsymbol{Z}_2, ...,\boldsymbol{Z}_R \in \mathbb{R}^{N \times K}$ that have been \textit{column-wise normalized}, the objective is minimizing the $L_2$ distance between each orthogonal features and their mean centers:
\begin{equation}
\begin{split}
\arg \min &  \sum_{r=1}^R  ||\boldsymbol{Z}_{r} - \Bar{\boldsymbol{Z}}||_F^2,   \\
\mathrm{ s.t. }\ \  \boldsymbol{Z}_r^T \boldsymbol{Z}_r &= \boldsymbol{I}, \forall \ r \in \{1,2, \ldots, R\},
\label{eq:cca}
\end{split}
\end{equation}
where $\Bar{\boldsymbol{Z}} =\frac{1}{R}\sum_{r=1}^R \Zv_r$ are the mean centers of the representations.

Eq.~\eqref{eq:cca} has orthogonality constraints thus cannot be optimized end-to-end with other objective functions, which could limit its effectiveness~\cite{soft_CCA}. Inspired by the soft decorrelation method proposed in~\cite{soft_CCA}, we \textit{formulate the orthogonality constraints as a loss function to achieve end-to-end learning.} The core idea is to approximate the full-batch covariance matrix at each training step by stochastic incremental learning and encourage sparsity in the off-diagonal elements of the approximated covariance matrix using $L_1$ regularization. Consider the mini-batch representation $\boldsymbol{Z}_{B} \in \mathbb{R}^{B \times K}$ of size $B$ which has been batch normalized~\cite{DBLP:bn}. At $t$-th training step, we compute the mini-batch covariance matrix $C_B^t = \frac{1}{B-1}\boldsymbol{Z}_B^T\boldsymbol{Z}_B$ and an accumulative covariance matrix over each mini batch as:
\begin{equation}
\boldsymbol{C}^t_{accu} = \alpha \boldsymbol{C}^{t-1}_{accu} +  \boldsymbol{C}^t_B,    
\end{equation}
where $\alpha \in [0,1)$ is a forgetting/decay rate, and $\boldsymbol{C}^{0}_{accu}$ is initialized as all-zero matrix. As such, the full-batch covariance matrix $\boldsymbol{C}^t = \boldsymbol{Z}^T\boldsymbol{Z}$ can be approximated by $\hat{\boldsymbol{C}^t}$ as:
\begin{equation}
    \hat{\boldsymbol{C}^t} = \frac{\boldsymbol{C}^t_{accu}}{c^t},
    \label{eq:C_appx}
\end{equation}
where $c^t = \alpha c^{t-1} + 1$ is a normalizing factor with $c^0 = 0$.

Given the approximate full-batch covariance matrix $\hat{\boldsymbol{C}^t}$ in Eq.~\eqref{eq:C_appx}, the orthogonality constraints $\boldsymbol{Z}^T\boldsymbol{Z}=\boldsymbol{I}$ can be achieved in a soft procedure by minimizing an $L_1$ loss in the off-diagonal elements to penalize the correlation. Denote the element of $\hat{\boldsymbol{C}^t}$ at entry $(i, j)$ as $\phi^t_{i,j}$. The soft orthogonality loss is defined as: 
\begin{equation}
    \mathcal{L}_{S}(\boldsymbol{Z}) = \sum_{i=1}^{K} \sum_{j = i + 1}^{K} |\phi^t_{i,j}|.
    \label{eq:soft_decorrelation_loss}
\end{equation}

 Based on Eq.~\eqref{eq:cca} and \eqref{eq:soft_decorrelation_loss}, we define our multi-scale alignment loss on top of  both $\boldsymbol{Z}^{\prime}$ and $\boldsymbol{Z}^{\prime\prime}$ as:
 \begin{equation}
     \mathcal{L}_A = \sum_{\boldsymbol{Z} \in \{\boldsymbol{Z}^{\prime}, \boldsymbol{Z}^{\prime\prime}\}}
     \bigg(  \sum_{r=1}^{R} ||\boldsymbol{Z}_{r} - \Bar{\boldsymbol{Z}}||_F^2 + \lambda_S \sum_{r=1}^{R} \mathcal{L}_S (\boldsymbol{Z}_{r}) \bigg),
 \end{equation}
where $\lambda_S$ controls the importance of the soft orthogonality loss. 

It is noteworthy that the idea behind $\mathcal{L}_A$ of aligning multi-scale information can be a general scheme for modeling MTS, which is worth further exploration in the future.     

\subsection{\rev{Summary and Complexity Analysis}}
\noindent
\textbf{Summary.}
Based on the discussion in Section~\ref{sec:multi-grained-contrasting} and Section~\ref{sec:multi-scale alignment}, the total loss of the proposed CSL is defined as: 
\begin{equation}
\mathcal{L} = \mathcal{L}_M + \lambda\mathcal{L}_A = \bigg( \mathcal{L}_C + \sum_{r=1}^{R}\mathcal{L}_{F,r} \bigg) + \lambda\mathcal{L}_A,\label{eq:total_loss}
\end{equation}
where $\lambda$ controls the importance of multi-scale alignment. 

The encoder (i.e., ST) $f(\boldsymbol{x}; \boldsymbol{\theta})$ is \textit{unsupervisedly} trained by minimizing the loss $\mathcal{L}$ using the popular back-propogation algorithm~\cite{chauvin2013backpropagation}, where $\boldsymbol{\theta}$ denotes trainable parameters which are updated within each mini batch.   The learned encoder maps MTS into latent representation as $\zv_i = f(\boldsymbol{x}_i; \boldsymbol{\theta})$, and $\zv_i$ is used for downstream tasks. 
\eat{\footnote{suggest to add some analysis or discussion of the proposed approach}}

\noindent
\rev{\textbf{Complexity analysis.} All data augmentation methods used in CSL take $O(BTD)$ time for MTS samples in a mini batch of size $B$~\cite{tsaug}. The encoder ST takes $O(BL_s(T-L_s+1)DD_{repr})$ time for embedding the time series into representations, where $L_s$ is the shapelet length. Recall that $RK = D_{repr}$. Therefore,  both $\mathcal{L}_C$ and $\sum_{r=1}^R\mathcal{L}_{F,r}$ can be computed in $O(B^2D_{repr})$ time and computing $\sum_{r=1}^R||\Zv_r - \Bar{\Zv}||_F^2$ takes $O(BD_{repr})$ time. The computation of $\mathcal{L}_S(\Zv_r)$ takes $O(B^2K)$ time dominated by computing $C_B^t = \frac{1}{B-1}\boldsymbol{Z}_B^T\boldsymbol{Z}_B$. Since each training epoch has $\lfloor\frac{N}{B}\rfloor$ batches, the total time complexity for training CSL is $O(NTD + NL_s(T-L_s+1)DD_{repr}+NBD_{repr})$. Considering that $B$ and $D_{repr}$ are constants and we set $L_s$ to be proportional to $T$, the time complexity can be simplified as $O(NT^2D)$. Similarly, the space complexity for the CSL training algorithm is $O(TD)$.}

\rev{We compare the complexity of CSL to that of the advanced URL baselines in Table~\ref{tab:complexity_analysis}. CSL is theoretically more scalable than TS2Vec and TST when $D >> T$, otherwise they have the same time complexity. CSL also has less space complexity than TS2Vec and TST. Compared to TS-TCC, the time complexity of CSL is the same or somewhat greater according to the relation between $T$ and $D$, but CSL has less space complexity. T-Loss and TNC are more scalable in both time and space. However, the two methods rely on many sequential operations which can not be accelerated by GPUs.  The experimental results in Section~\ref{sec:exp:time} show that they run much slower than CSL, TS2Vec, TST and TS-TCC with a considerably large input scale. Moreover, we show that our CSL, though primarily designed for improving the representation quality, also has faster training speed for real-world tasks, saying that it can achieve better performance with equal or less training time. }

\begin{table}[t]
    \centering
    \caption{\rev{Time and space complexity for CSL and the URL baselines.}}
    \vspace{-2.5ex}
    \resizebox{\linewidth}{!}{\begin{tabular}{lcccccc}
\toprule
\textbf{Complexity}	&  TS2Vec & T-Loss & TNC & TS-TCC & TST & \textbf{CSL} \\
\midrule

Time & $O(NT(T + D^2))$ & $O(NTD)$ & $O(NTD)$ & $O(NT(T + D))$ & $O(NT(T + D^2))$ & $O(NT^2D)$ \\

Space & $O(TD + T^2)$ & $O(D)$ & $O(D)$ & $O(TD + T^2)$ & $O(TD + T^2)$ & $O(TD)$ \\

\bottomrule
    \end{tabular}}
    \label{tab:complexity_analysis}
    \vspace{-3ex}
\end{table}

\section{Experiments}
\label{Sec: Experiments}
\subsection{Experimental Setup}\label{sec:setup}
{We conduct extensive experiments using total 34 real-world datasets to assess the representation quality of CSL. Three main tasks are investigated including the supervised classification task and the unsupervised clustering and anomaly detection tasks.} Note that URL considers the MTS representation at the segment level, thus we work on\eat{series-level} {segment-level} anomaly detection (rather than observation-level~\cite{SMD-KDD19,ASD-KDD21}). In specific, we consider series at each sliding window $\xv_i[t,w], t=1,2,...,N - w + 1$ as an anomaly if it contains at least one anomalous observation. We train the popular SVM, K-means, and Isolation Forest on top of the learned representations to solve the three tasks respectively.   
We describe the datasets, baselines, implementations and evaluation metrics as follows.

\begin{table}[t]
    \centering
    \caption{Statistics of the 30 UEA datasets. All datasets are used for classification evaluation and the 12 subsets marked by $^*$ are used for clustering evaluation following~\cite{MUSLA}.}
    \vspace{-2ex}
    \resizebox{\linewidth}{!}{\begin{tabular}{lccccc}
\toprule
\textbf{Dataset}	& \# Train &	\# Test &	\# Dim &	Length &	\# Class \\
\midrule

ArticularyWordRecognition (AW)$^*$ & 275 & 300 & 9 & 144 & 25 \\

AtrialFibrillation (AF)$^*$ & 15 & 15 & 2 & 640 & 3 \\

BasicMotions (BM)$^*$ & 40 & 40 & 6 & 100 & 4 \\

CharacterTrajectories (CT) & 1422 & 1436 & 3 & 182 & 20 \\

Cricket (Cr) & 108 & 72 & 6 & 1197 & 12 \\

DuckDuckGeese (DD) & 50 & 50 & 1345 & 270 & 5 \\

EigenWorms (EW) & 128 & 131 & 6 & 17984 & 5 \\

Epilepsy (Ep)$^*$ & 137 & 138 & 3 & 206 & 4 \\

EthanolConcentration (EC) & 261 & 263 & 3 & 1751 & 4 \\

ERing (ER)$^*$ & 30 & 270 & 4 & 65 & 6 \\

FaceDetection (FD) & 5890 & 3524 & 144 & 62 & 2 \\

FingerMovements (FM) & 316 & 100 & 28 & 50 & 2 \\

HandMovementDirection (HM)$^*$ & 160 & 74 & 10 & 400 & \textit{4} \\

Handwriting (Ha) & 150 & 850 & 3 & 152 & 26 \\

Heartbeat (He) & 204 & 205 & 61 & 405 & 2 \\

InsectWingbeat (IW) & 30000 & 20000 & 200 & 30 & 10 \\

JapaneseVowels (JV) & 270 & 370 & 12 & 29 & 9 \\

Libras (Li)$^*$ & 180 & 180 & 2 & 45 & 15 \\

LSST (LS)& 2459 & 2466 & 6 & 36 & 14 \\

MotorImagery (MI) & 278 & 100 & 64 & 3000 & 2 \\

NATOPS (NA)$^*$ & 180 & 180 & 24 & 51 & 6 \\

PenDigits (PD)$^*$ & 7494 & 3498 & 2 & 8 & 10 \\

PEMS-SF (PE)$^*$ & 267 & 173 & 963 & 144 & 7 \\

PhonemeSpectra (PS) & 3315 & 3353 & 11 & 217 & 39 \\

RacketSports (RS) & 151 & 152 & 6 & 30 & 4 \\

SelfRegulationSCP1 (SR1) & 268 & 293 & 6 & 896 & 2 \\

SelfRegulationSCP2 (SR2) & 200 & 180 & 7 & 1152 & 2 \\

SpokenArabicDigits (SA) & 6599 & 2199 & 13 & 93 & 10 \\

StandWalkJump (SW)$^*$ & 12 & 15 & 4 & 2500 & 3 \\

UWaveGestureLibrary (UW)$^*$ & 120 & 320 & 3 & 315 & 8 \\

\bottomrule
    \end{tabular}}
    \label{tab:uea_statistics}
    \vspace{-2ex}
\end{table}

\begin{table}[t]
    \centering
    
    \caption{Statistics of used anomaly detection datasets.}
    \vspace{-2.5ex}
    \resizebox{.88\linewidth}{!}{\begin{tabular}{lccccc}
\toprule
\textbf{Dataset}	&  \# Entity  & \# Dim & Train length & Test length & Anomaly ratio (\%) \\
\midrule

SMAP & 55 & 25 & 135183 & 427617 & 13.13 \\

MSL & 27 & 55 & 58317 & 73729 & 10.72 \\

SMD & 12 & 38 & 304168 & 304174 & 5.84 \\

ASD & 12 & 19 & 102331 & 51840 & 4.61 \\

\bottomrule
    \end{tabular}}
    \label{tab:ad_statistics}
    \vspace{-2ex}
\end{table}

\noindent
\textbf{Datasets.}
We use 34 MTS datasets with various sample size, dimension, series length, number of classes and application scenario to evaluate the representation quality on the three downstream tasks. We use the default train/test split for all datasets where only the training data are used for learning the encoder and task-specific models. The datasets used for each task are present below.  

\textit{(1) \underline{Classification.}} To benchmark the result, we evaluate the performance of MTS classification on all 30 datasets of the popular UEA  archive~\cite{UEA}. These data are collected from various domains, e.g., human action recognition, Electrocardiography monitoring and audio classification. The dataset statistics is present in Table~\ref{tab:uea_statistics}.

\textit{(2) \underline{Clustering.}} Following a recent work of multivariate time series clustering~\cite{MUSLA}, we evaluate the clustering performance using 12 UEA subsets which are highly heterogenerous in train/test size, length, and the number of dimensions and classes. The statistics of these 12 datasets are shown in Table~\ref{tab:uea_statistics} (marked by $^*$). 

\textit{(3) \underline{Anomaly Detection.}}  Four recently published datasets collected from several challenging real-world applications are used for anomaly detection. Soil Moisture Active Passive satellite (SMAP) and Mars Science Laboratory rover (MSL) are two spacecraft anomaly detection datasets from NASA~\cite{SMAP&MSL-KDD18}. Server Machine Dataset (SMD) is a 5-week-long dataset collected by~\cite{SMD-KDD19} from a large Internet company.  Application Server Dataset (ASD) is a 45-day-long MTS charactering the status of the servers recently collected by~\cite{ASD-KDD21}. Following~\cite{ASD-KDD21}, for SMD, we use the 12 entities that do not suffer concept drift for evaluation. Table~\ref{tab:ad_statistics} shows the dataset statistics.














\noindent
\textbf{Baselines.}
We use \rev{21} baselines for comparison, which are categorized into two groups:

    \textit{(1) \underline{URL methods.}} We compare our CSL with 5 URL baselines specially designed for time series, including TS2Vec~\cite{ts2vec}, T-Loss~\cite{T-Loss}, TNC~\cite{TNC}, TS-TCC~\cite{TS-TCC}, and TST~\cite{TST}. All URL competitors are evaluated in the same way as CSL for a fair comparison. More details of these methods are discussed in Section~\ref{sec:related_work}. 
    
    \textit{(2) \underline{Task-specific methods.} }We also include baselines tailored for downstream tasks. We select outstanding approaches \eat{of diverse types} for classification,  containing the most popular baseline DTWD which adopts the one-nearest-neighbor classifier with dynamic time warping as the distance metric~\cite{UEA} and five supervised techniques, including the RNN-based MLSTM-FCNs~\cite{MLSTM-FCNs}, the attentional prototype-based TapNet~\cite{Tapnet}, the shapelet-based ShapeNet~\cite{Shapenet}, and the CNN-based OSCNN~\cite{OSCNN} and DSN~\cite{DSN}. To avoid an unfair comparison, we let outside
the ensemble methods like~\cite{hive-cote}. \rev{Recall that the supervised classification methods use the true labels to learn the features, which is benchmarked against the data augmentation or sampling in URL. Thus, the comparison is fair without further applying the data augmentation methods of CSL on the baseline approaches.  }

\rev{We consider six advanced} clustering baselines including the dimension reduction-based MC2PCA~\cite{mc2pca} and TCK~\cite{TCK}, the distance-based m-kAVG+ED and m-kDBA~\cite{mkavg-mkdba-mksc}, the deep learning-based DeTSEC~\cite{DeTSEC}, and the shapelet-based MUSLA~\cite{MUSLA}. \rev{In addition, we design   ShapeNet-Clustering (SN-C), an adaption of the classification baseline ShapeNet which is also based on the shapelets. In SN-C, we dismiss the supervised feature selection of ShapeNet and use K-means rather than SVM upon the features for clustering.} 

Since no evaluation is reported on the anomaly detection datasets under the segment-level setting, we develop 2 baselines on top of the raw MTS using also Isolation Forest for a fair comparison.  The models take the observations either at each timestamp (denoted as IF-p) or within each sliding window (denoted as IF-s) as the input. \rev{Similarly, we also adapt ShapeNet for anomaly detection (SN-AD) by dismissing the supervised feature selection of ShapeNet and using Isolation Forest upon the shapelet-transformed features.} 

\begin{figure*}
    \centering
\subcaptionbox{\centering TS2Vec.
    \label{subfig:ts2vec-repr}}
{\includegraphics[width=.125\linewidth]{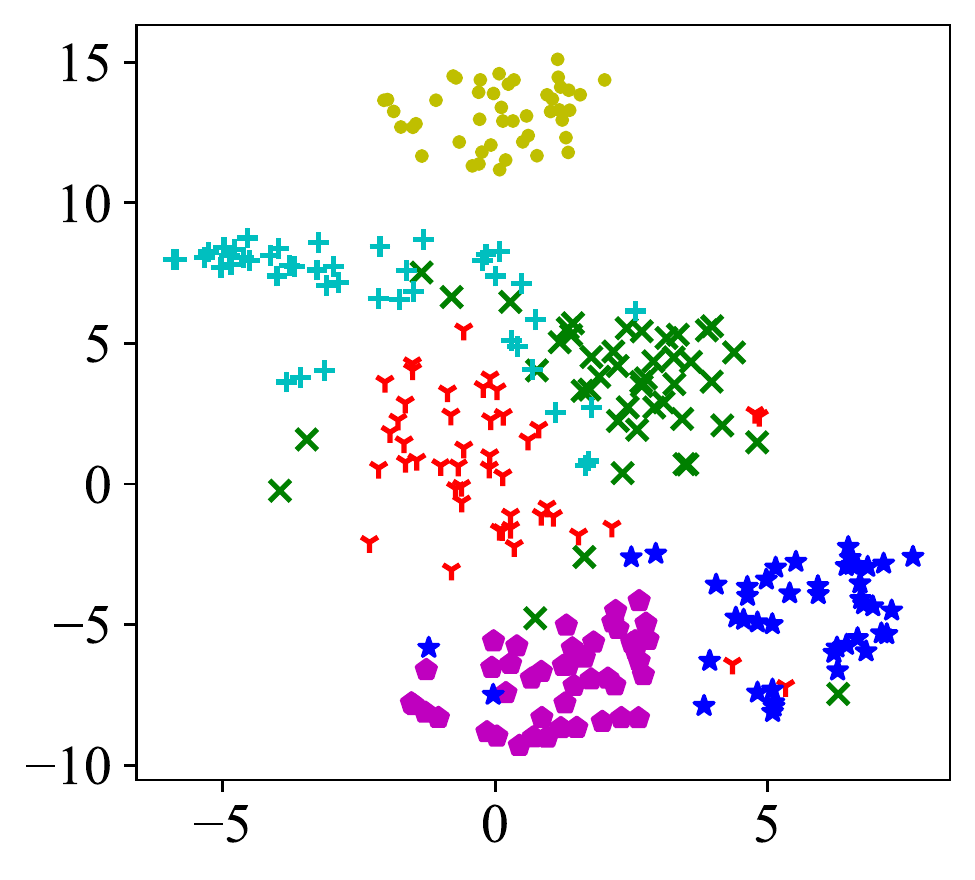}\vspace{-1ex}}
\subcaptionbox{\centering T-Loss.
    \label{subfig:tloss-repr}}
{\includegraphics[width=.125\linewidth]{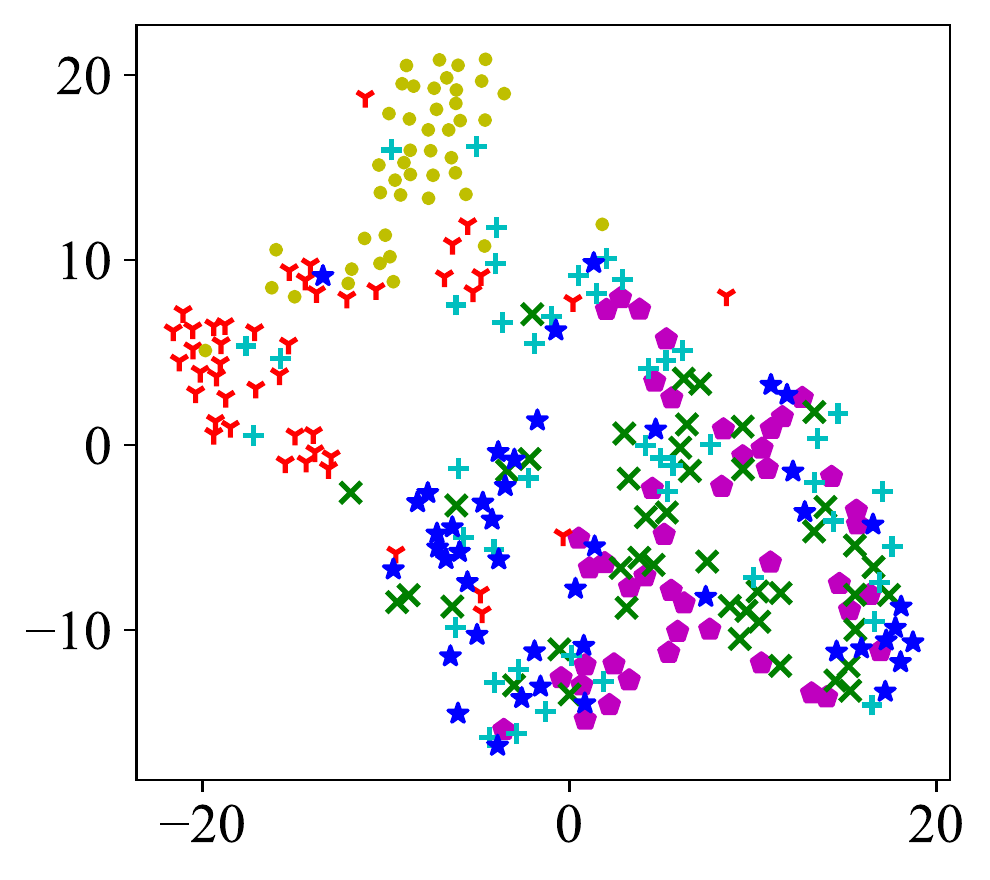}\vspace{-1ex}}
\subcaptionbox{\centering TNC.
    \label{subfig:tnc-repr}}
{\includegraphics[width=.125\linewidth]{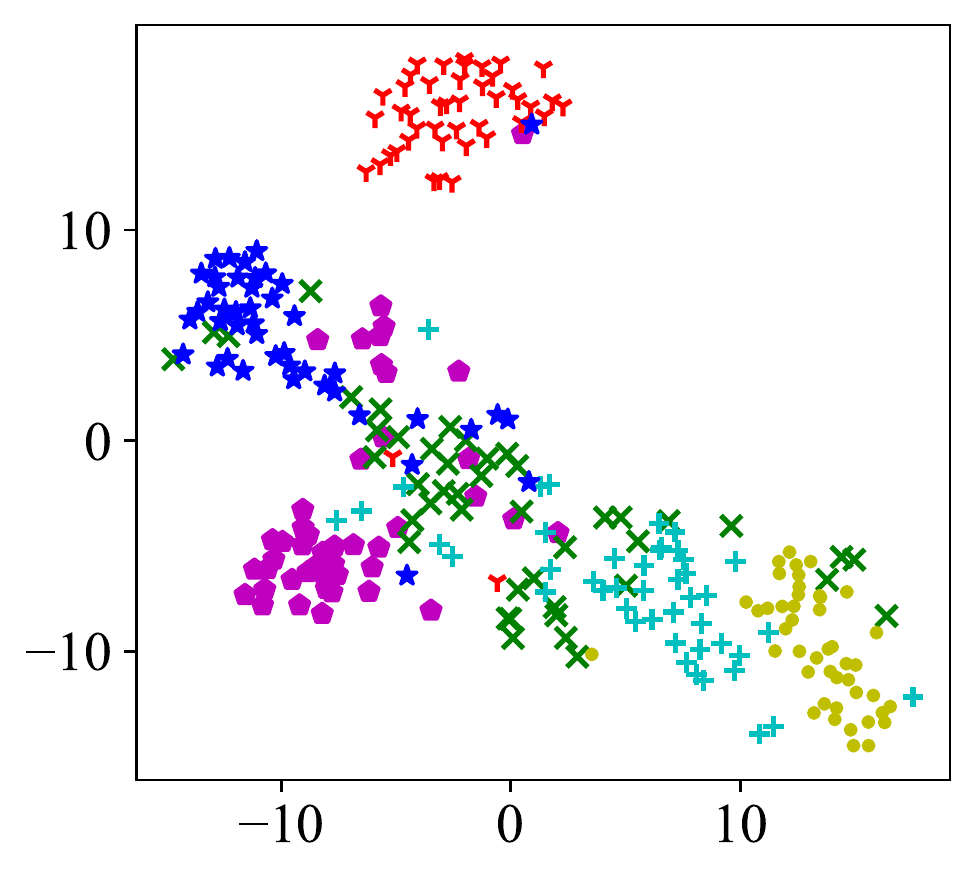}\vspace{-1ex}}
\subcaptionbox{\centering TS-TCC.
    \label{subfig:tstcc-repr}}
{\includegraphics[width=.125\linewidth]{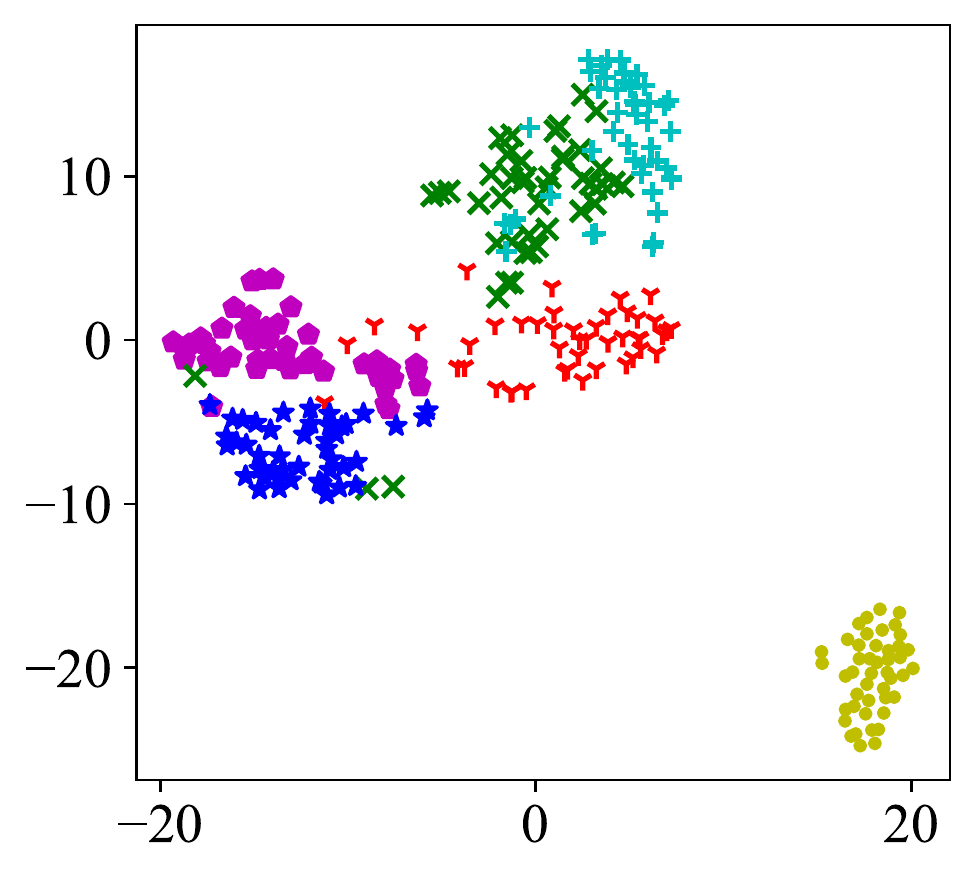}\vspace{-1ex}}
\subcaptionbox{\centering TST.
    \label{subfig:tst-repr}}
{\includegraphics[width=.125\linewidth]{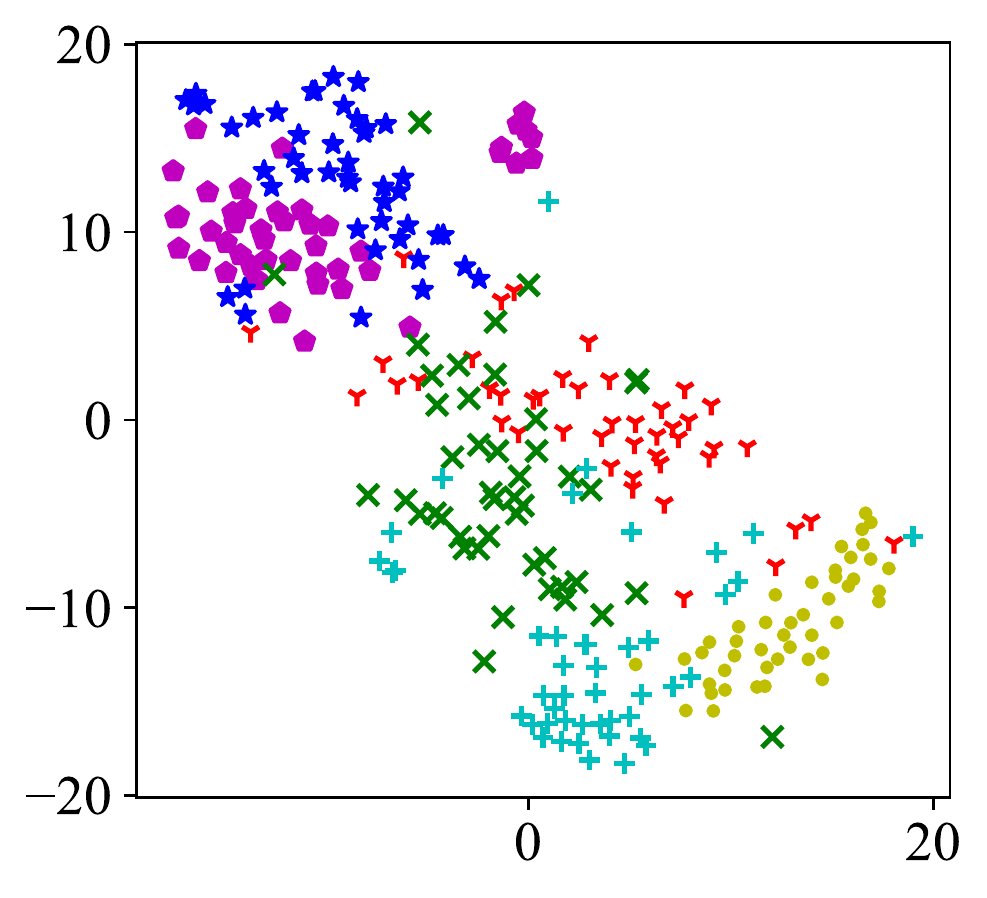}\vspace{-1ex}}
\subcaptionbox{\centering ShapeNet.
    \label{subfig:ShapeNet-repr}}
{\includegraphics[width=.125\linewidth]{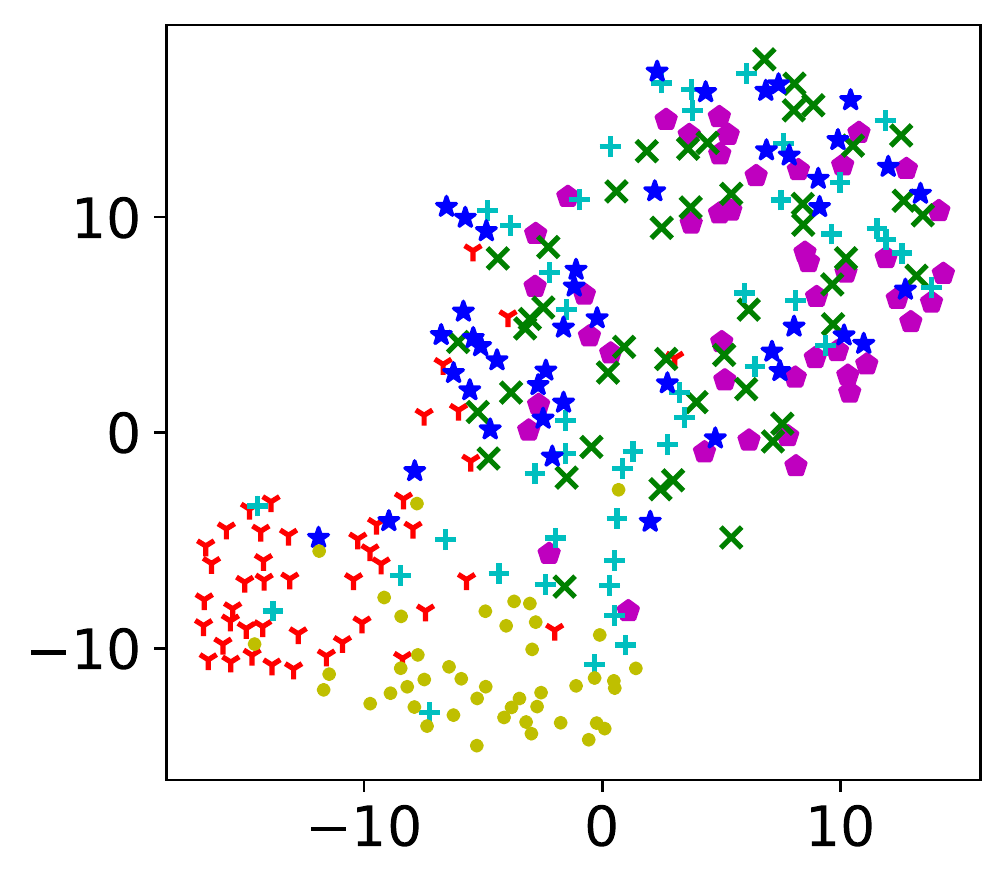}\vspace{-1ex}}
\subcaptionbox{\centering CSL.
    \label{subfig:csl-repr}}
{\includegraphics[width=.125\linewidth]{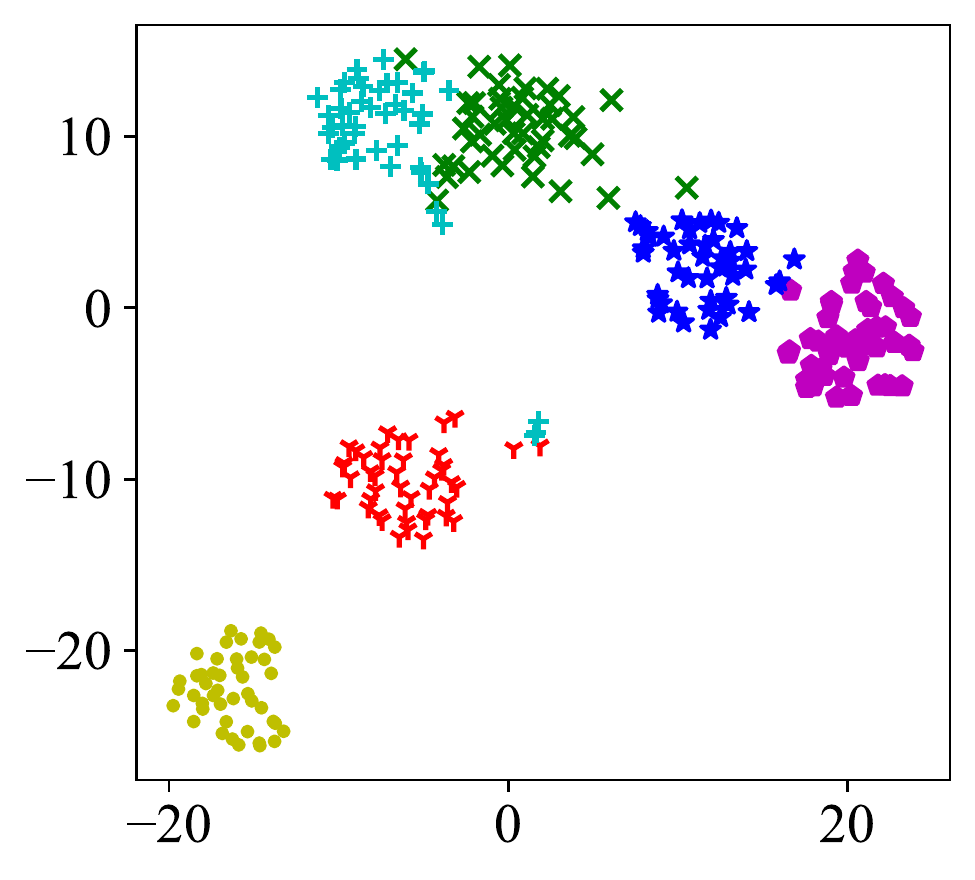}\vspace{-1ex}}
\vspace{-2.5ex}
    \caption{\rev{Two-dimensional t-SNE~\cite{tSNE} visualization of the unsupervised learned representation for ERing test set. Classes are distinguishable using their respective marker shapes and colors.}}
    \label{fig:repr}
    \vspace{-1ex}
\end{figure*}

\begin{table*}[t]
    \caption{Performance comparison on MTS classification. The best results among URL methods are highlighted in bold and $\dag$ indicates the best among all competitors. The underlined value indicates significant difference under a statistical level of 0.05.}
    \vspace{-2.5ex}
    \label{tab:classification}
    \resizebox{.77\textwidth}{!}{
    \begin{tabular}{lcccccc|cccccc}
    \toprule
    \multirow{2}{*}{\textbf{Dataset}} & \multicolumn{6}{c|}{\textbf{Tailored Classification Approaches}} & \multicolumn{6}{c}{\textbf{Unsupervised Representation Learning + Classifier}} \\
    \cline{2-13}

\rule{0pt}{3ex}  & DTWD	& MLSTM-FCNs  &TapNet &ShapeNet	& OSCNN	 & DSN        &TS2Vec  &T-Loss &TNC	 &TS-TCC &TST & \textbf{CSL}  \\
    \midrule
    	
\eat{ArticularyWordRecognition} AW   & 0.987   & 0.973        & 0.987    & 0.987         & 0.988        & 0.984         & 0.987                 & 0.943          & 0.973        & 0.953            & 0.977       & \textbf{0.990}$^\dag$   \\
\eat{AtrialFibrillation} AF   & 0.200     & 0.267        & 0.333    & 0.400	        & 0.233	       & 0.067         & 0.200	               & 0.133	        & 0.133	       & 0.267	          & 0.067	      & \textbf{0.533}$^\dag$  \\
\eat{BasicMotions} BM   & 0.975	  & 0.950	    & 1.000$^\dag$ & 1.000$^\dag$      & 1.000$^\dag$     & 1.000$^\dag$      & 0.975	             &\textbf{1.000}$^\dag$	& 0.975	       & \textbf{1.000}$^\dag$& 0.975	      & \textbf{1.000}$^\dag$	   \\
\eat{CharacterTrajectories} CT  & 0.989	  & 0.985	     & 0.997	& 0.980	        & 0.998$^\dag$ & 0.994         & \textbf{0.995}        & 0.993	        & 0.967	       & 0.985	          & 0.975	      & 0.991           \\
\eat{Cricket} Cr   &1.000$^\dag$ & 0.917      	 & 0.958	& 0.986	        & 0.993	       & 0.989         & 0.972	               & 0.972	        & 0.958	       & 0.917	      &\textbf{1.000}$^\dag$   & 0.994           \\
\eat{DuckDuckGeese} DD   & 0.600     & 0.675	     & 0.575	& 0.725$^\dag$  & 0.540	       & 0.568         & \textbf{0.680}         & 0.650	        & 0.460	       & 0.380	          & 0.620	      & 0.380            \\
\eat{EigenWorms} EW  & 0.618	  & 0.504	   	 & 0.489	& 0.878$^\dag$	& 0.414        & 0.391         & \textbf{0.847}        & 0.840	        & 0.840	       & 0.779	          & 0.748	      & 0.779           \\
\eat{Epilepsy} Ep   & 0.964	  & 0.761	    	 & 0.971	& 0.987	        & 0.980	       & 0.999$^\dag$  & 0.964	               & 0.971	        & 0.957	       & 0.957	          & 0.949	      & \textbf{0.986}  \\
\eat{ERing} ER   & 0.133	  & 0.133	    	 & 0.133	& 0.133	        & 0.882	       & 0.922         & 0.874	               & 0.133	        & 0.852	       & 0.904	          & 0.874	      & \textbf{0.967}$^\dag$  \\
\eat{EthanolConcentration} EC  & 0.323	  & 0.373	   	 & 0.323	& 0.312	        & 0.241	       & 0.245         & 0.308	               & 0.205	        & 0.297	       & 0.285	          & 0.262	      & \textbf{0.498}$^\dag$  \\
\eat{FaceDetection} FD  & 0.529	  & 0.545	    	 & 0.556	& 0.602	        & 0.575	       & 0.635$^\dag$  & 0.501	               & 0.513	        & 0.536	       & 0.544	          & 0.534	      & \textbf{0.593}  \\
\eat{FingerMovements} FM  & 0.530	  & 0.580	    	 & 0.530	    & 0.580	        & 0.568	       & 0.492         & 0.480	               & 0.580	        & 0.470	       & 0.460	          & 0.560	      & \textbf{0.59}$^\dag$   \\
\eat{HandMovementDirection} HM  & 0.231	  & 0.365	   	 & 0.378	& 0.338	        & 0.443$^\dag$ & 0.373         & 0.338	               & 0.351	        & 0.324	       & 0.243	          & 0.243	      & \textbf{0.432}  \\
\eat{Handwriting} Ha  & 0.286	  & 0.286	    	 & 0.357	& 0.451	        & 0.668$^\dag$ & 0.337         & 0.515	               & 0.451	        & 0.249	       & 0.498	          & 0.225	      & \textbf{0.533}  \\
\eat{Heartbeat} He  & 0.717	  & 0.663	    	 & 0.751	& 0.756	        & 0.489	       & 0.783$^\dag$  & 0.683	               & 0.741	        & 0.746	       &\textbf{0.751}	  & 0.746	      & 0.722           \\
\eat{InsectWingbeat} IW  & N/A	      & 0.167	   	 & 0.208	& 0.250	        & 0.667$^\dag$ & 0.386         & 0.466                 & 0.156	        &\textbf{0.469}& 0.264	          & 0.105	      & 0.256           \\  
\eat{JapaneseVowels} JV  & 0.949	  & 0.976	    	 & 0.965	& 0.984	        & 0.991$^\dag$ & 0.987         & 0.984	               & \textbf{0.989} & 0.978	       & 0.930	          & 0.978	      & 0.919           \\
\eat{Libras} Li & 0.870	  & 0.856	   	 & 0.850	    & 0.856	        & 0.950	       & 0.964$^\dag$  & 0.867	               & 0.883	        & 0.817	       & 0.822	          & 0.656	      & \textbf{0.906}  \\
\eat{LSST} LS  & 0.551	  & 0.373	    	 & 0.568    & 0.590	        & 0.413	       & 0.603         & 0.537	               & 0.509	        & 0.595	       & 0.474	          & 0.408	      & \textbf{0.617}$^\dag$  \\
\eat{MotorImagery} MI & 0.500	  & 0.510	    	 & 0.590	    &0.610$^\dag$    & 0.535	       & 0.574         & 0.510	               & 0.580	        & 0.500	       & 0.610$^\dag$	  & 0.500	      & \textbf{0.610}$^\dag$   \\
\eat{NATOPS} NA  & 0.883	  & 0.889	    	 & 0.939	& 0.883	        & 0.968	       & 0.978$^\dag$  & \textbf{0.928}        & 0.917          & 0.911	       & 0.822	          & 0.850	      & 0.878           \\
\eat{PEMS-SF} PE  & 0.711	  & 0.699	   	 & 0.751	& 0.751	        & 0.760	       & 0.801         & 0.682	               & 0.676	        & 0.699	       & 0.734	          & 0.740	      & \textbf{0.827}$^\dag$  \\
\eat{PenDigits} PD  & 0.977	  & 0.978	     & 0.980	    & 0.977	        & 0.986	       & 0.987         & 0.989	               & 0.981	        & 0.979	       & 0.974	          & 0.56	      & \textbf{0.990}$^\dag$   \\
\eat{PhonemeSpectra} PS  & 0.151	  & 0.110	    	 & 0.175	& 0.298	        & 0.299$^\dag$ & 0.320          & 0.233	               & 0.222	        & 0.207	       & 0.252	          & 0.085	      & \textbf{0.255}  \\
\eat{RacketSports} RS  & 0.803	  & 0.803	    	 & 0.868	& 0.882$^\dag$	& 0.877	       & 0.862         & 0.855	               & 0.855	        & 0.776	       & 0.816	          & 0.809	      & \textbf{0.882}$^\dag$  \\
\eat{SelfRegulationSCP1} SR1  & 0.775	  &0.874$^\dag$		 & 0.652	& 0.782	        & 0.835	       & 0.717         & 0.812	               & 0.843	        & 0.799	       & 0.823	          & 0.754	      & \textbf{0.846}  \\
\eat{SelfRegulationSCP2} SR2  & 0.539	  & 0.472	    	 & 0.550	    & 0.578$^\dag$	& 0.532	       & 0.464         & \textbf{0.578}$^\dag$ & 0.539	        & 0.550	       & 0.533	          & 0.550	      & 0.494           \\
\eat{SpokenArabicDigits} SA  & 0.963	  & 0.990	    	 & 0.983	& 0.975	        & 0.997$^\dag$ & 0.991         & 0.988	               & 0.905	        & 0.934	       & 0.970	          & 0.923	      & \textbf{0.990}   \\
\eat{StandWalkJump} SW  & 0.200	  & 0.067      	 & 0.400	    & 0.533	        & 0.383	       & 0.387         & 0.467	               & 0.333	        & 0.400	       & 0.333	          & 0.267	      & \textbf{0.667}$^\dag$  \\
\eat{UWaveGestureLibrary} UW  & 0.903	  & 0.891	      & 0.894	& 0.906	        & 0.927$^\dag$ & 0.916         & 0.906	               & 0.875	        & 0.759	       & 0.753	          & 0.575	      & \textbf{0.922}  \\
 
\hline
\textbf{Avg} (excl. IW) & 0.650	  & 0.637	    	 & 0.673	& 0.714	        & 0.706	       & 0.701         & 0.712  & 0.675 & 0.677 & 0.682 & 0.635  & \textbf{0.751}$^\dag$  \\
\textbf{Avg} (incl. IW) & N/A & 0.621  & 0.657 & 0.699 &
       0.704 & 0.691 & 0.704 & 0.658 & 0.670 &
       0.668 & 0.617 & \textbf{0.735}$^\dag$ \\
\hline
\makecell[l]{\textbf{AR} (URL)} & / & / & / & / & / & /  & 2.82 & 3.52 & 4.02  & 3.87 & 4.70 & \textbf{2.08} \\
\makecell[l]{\textbf{AR} (All)} & 
8.17 & 8.10 & 6.00 &  4.68 & 4.53 &
   4.98 & 5.87 & 6.97 & 8.10  & 7.82 &
       9.02 & \textbf{3.77}$^\dag$ \\
\hline
\makecell[l]{\eat{\textbf{1 vs. 1 Compasion} \\} \textbf{W/T/L}} &
25/0/5 & 25/1/4  & 23/1/6 & 18/3/9 & 17/1/12 & 18/1/11 & 23/0/7 & 22/1/7 & 23/0/7 & 22/4/4 & 25/0/5 & / \\
\hline
\textbf{p-val} & \underline{0.0003} & \underline{0.0003}  & \underline{0.0038} & 0.1075 & 0.4174 & 0.2177 & \underline{0.0316} & \underline{0.0058} & \underline{0.0066} & \underline{0.0002} & \underline{0.0002} & / \\
    \bottomrule
    \end{tabular}
    }\vspace{-1ex}
\end{table*}

\noindent \textbf{Implementations.}
We implement the CSL model using PyTorch 1.10.2 and run all experiments on a Ubuntu machine with Tesla V100 GPU. The SVM, K-means, and Isolation Forest are implemented using Scikit-learn 1.1.1 and the data augmentation methods are implemented using tsaug~\cite{tsaug} with default parameters. Most of the hyper-parameters of CSL are set to fixed values for all experiments without tuning. We adopt SGD optimizer to learn the ST encoder. The learning rate is set to 0.01. Batchnorm is applied after the encoding. We set $\alpha = 0.5$ in soft orthogonality and $\lambda = 0.01$, $\lambda_S = 1$ in loss functions. The batch size is set to 8 for all UEA datasets and 256 for the anomaly detection datasets. The temperature $\tau$ is selected from $\{0.1, 0.01, 0.001\}$ by cross validation for the UEA datasets and is fixed to 0.1 for the anomaly detection datasets.   Following previous works~\cite{ts2vec,MUSLA}, the embedding dimension is fixed to $D_{repr} = 320$ for classification and is chosen from $\{80, 240, 320\}$ for clustering for a fair comparison. On anomaly detection tasks, we set $D_{repr}$ to 240, 320, 48, 32 for SMAP, MSL, SMD, and ASD respectively.

\begin{table*}[h]
    \centering
    \caption{Performance comparison on MTS clustering. The best results among URL methods are highlighted in bold, and $\dag$ indicates the best among all competitors. The underlined value indicates significant difference under a statistical level of 0.05.}
    \vspace{-2.5ex}
    \label{tab:clustering}
    \resizebox{.86\linewidth}{!}{
    \begin{tabular}{lcccccccc|cccccc}
    \toprule
    \multirow{2}{*}{\textbf{Dataset}} & \multirow{2}{*}{\textbf{Metric}} & \multicolumn{7}{c|}{\textbf{Tailored Clustering Approaches}} & \multicolumn{6}{c}{\textbf{Unsupervised Representation Learning + Clustering}} \\
    \cline{3-15}&
     \rule{0pt}{3ex} &  MC2PCA & TCK & m-kAVG+ED &	 m-kDBA &	 DeTSEC		& MUSLA  & \rev{SN-C} & TS2Vec	& T-Loss & TNC	& TS-TCC	& TST & \textbf{CSL}\\
    \midrule
\multirow{2}{*}{\eat{ArticularyWordRecognition}AW} & RI	& 0.989 & 0.973 &	0.952 &	0.934 &	0.972  &	0.977 & \rev{0.936} &	0.980 &	0.975 &	0.938 &	0.946 &	0.978 &	\textbf{0.990}$^\dag$     \\
                                           & NMI & 0.934 &	0.873 &	0.834 &	0.741 &	0.792  &	0.838  & \rev{0.492} & 0.880 &	0.842 &	0.565 &	0.621 &	0.866 &	\textbf{0.942}$^\dag$       \\
\multirow{2}{*}{\eat{AtrialFibrillation}AF} & RI	& 0.514 & 0.552 &  	0.705 &	0.686 &	0.629  &	0.724    &  \rev{0.410} & 0.465 &	0.469 &	0.518 &	0.469 &	0.444 &	\textbf{0.743}$^\dag$      \\
                                           & NMI & 0.514	& 0.191 &  	0.516 &	0.317 &	0.293  &	0.538 & \rev{0.213} & 0.080 &	0.149 &	0.147 &	0.164 &	0.171 &	\textbf{0.587}$^\dag$    \\
\multirow{2}{*}{\eat{BasicMotions}BM} & RI & 0.791	& 0.868 &  	0.772 &	0.749 &	0.717  &	1.000$^\dag$ & \rev{0.836} &	0.854 &	0.936 &	0.719 &	0.856 &	0.844 &	\textbf{1.000}$^\dag$      \\
                                           & NMI & 0.674	& 0.776 &  	0.543 &	0.639 &	0.800  &	1.000$^\dag$ & \rev{0.736} &	0.820 &	0.871 &	0.394 &	0.823 &	0.790 &	\textbf{1.000}$^\dag$    \\
\multirow{2}{*}{\eat{Epilepsy}Ep} & RI & 0.613	& 0.786 &  	0.768 &	0.777 &	0.840  &	0.816 & \rev{0.695} &	0.706 &	0.705 &	0.650 &	0.736 &	0.718 &	\textbf{0.873}$^\dag$       \\
                                           & NMI & 0.173	& 0.534 &  	0.409 &	0.471 &	0.346  &	0.601 & \rev{0.250} &	0.312 &	0.306 &	0.156 &	0.451 &	0.357 &	\textbf{0.705}$^\dag$     \\
\multirow{2}{*}{\eat{ERing}ER} & RI	& 0.756 & 0.772 &  	0.805 &	0.775 &	0.770  &	0.841 & \rev{0.734} &	0.925 &	0.885 &	0.764 &	0.821 &	0.867 &	\textbf{0.968}$^\dag$      \\
                                           & NMI & 0.336	& 0.399 &  	0.400 &	0.406 &	0.392  &	0.722 & \rev{0.349} &	0.775 &	0.672 &	0.346 &	0.478 &	0.594 &	\textbf{0.906}$^\dag$    \\
\multirow{2}{*}{\eat{HandMovementDirection}HM} & RI & 0.627	& 0.635 &  	0.697 & 0.685 &	0.628  &	0.719$^\dag$ & \rev{0.636} &	0.609 	& 0.599 & 0.613 &	0.608 &	0.607 &	\textbf{0.651}      \\
                                           & NMI & 0.067	& 0.103 &  	0.168 &	0.265 &	0.112  &	0.398$^\dag$ & \rev{0.125} &	0.044 &	0.034 &	0.051 &	0.053 &	0.039 &	\textbf{0.175}    \\
\multirow{2}{*}{\eat{Libras}Li} & RI & 0.892 	& 0.917 &  	0.911 &	0.913 &	0.907  &	0.941$^\dag$ & \rev{0.855} &	0.904 &	0.922 &	0.896 &	0.881 &	0.886 &	\textbf{0.941}$^\dag$      \\
                                           & NMI	& 0.577 & 0.620 &  	0.622 &	0.622 &	0.602  &	0.724 & \rev{0.353} &	0.542 & 0.654 &	0.464 &	0.373 &	0.400 &	\textbf{0.761}$^\dag$    \\
\multirow{2}{*}{\eat{NATOPS}NA} & RI & 0.882	& 0.833 &  	0.853 &	0.876 &	0.714  &	0.976$^\dag$ & \rev{0.754} &	0.817 &	0.836 &	0.700 &	0.792 &	0.809 &	\textbf{0.876}      \\
                                           & NMI & 0.698	& 0.679 &  	0.643 &	0.643 &	0.043  &	0.855$^\dag$ & \rev{0.313}	& 0.523 &	0.558 &	0.222 &	0.513 &	0.565 &	\textbf{0.657}    \\
\multirow{2}{*}{\eat{PEMS-SF}PE} & RI & 0.424	& 0.191 &  	0.817 &	0.755 &	0.806  &	0.892$^\dag$ & \rev{0.730} &	0.765 &	0.746 &	0.763 &	0.789 &	0.726 &	\textbf{0.858}      \\
                                           & NMI & 0.011	& 0.066 &  	0.491 &	0.402 &	0.425  &	0.614$^\dag$ & \rev{0.217} &	0.290 &	0.102 & 0.278 &	0.331 &	0.026 &	\textbf{0.537}    \\
\multirow{2}{*}{\eat{PenDigits}PD} & RI	& 0.929 & 0.922 &  	0.935 &	0.881 &	0.885  &	0.946 & \rev{0.845} &	0.941 &	0.936 &	0.873 &	0.857 &	0.818 &	\textbf{0.950}$^\dag$      \\
                                           & NMI & 0.713	& 0.693 &  	0.738 &	0.605 &	0.563  &	0.826$^\dag$ & \rev{0.273} &	0.776 & 0.749 &	0.537 &	0.339 &	0.090 & \textbf{0.822}    \\
\multirow{2}{*}{\eat{StandWalkJump}SW} & RI & 0.591	& 0.762 &  	0.733 &	0.695 &	0.733  &	0.771$^\dag$ & \rev{0.467} &	0.410 &	0.410 &	0.457 &	0.589 &	0.467 &	\textbf{0.724}      \\
                                           & NMI & 0.350	& 0.536 &  	0.559 &	0.466 &	0.556  &	0.609$^\dag$ & \rev{0.188} & 	0.213 &	0.213 &	0.193 &	0.187 &	0.248 &	\textbf{0.554}    \\
\multirow{2}{*}{\eat{UWaveGestureLibrary}UW} & RI & 0.883	& 0.913 &  	0.920 &	0.893 &	0.879  &	0.913 & \rev{0.795} &	0.865 &	0.893 &	0.817 &	0.796 &	0.779 &	\textbf{0.927}$^\dag$      \\
                                           & NMI & 0.570	& 0.710 &  	0.713 &	0.582 &	0.558  &	0.728 &	\rev{0.233} & 0.511 &	0.614 & 0.322 &	0.215 &	0.244 &	\textbf{0.731}$^\dag$    \\
\hline     
\multirow{2}{*}{\textbf{Avg}}             & RI	& 0.741 & 0.760 & 	0.822	& 0.801	   &0.790		&0.876$^\dag$ & \rev{0.724}	& 0.770	&0.776	&0.726	&0.762	&0.745	&\textbf{0.875}     \\
                                           & NMI & 0.468 	& 0.515 & 	0.553 &  0.513	&0.457	   &0.704$^\dag$ & \rev{0.312} & 0.480	 &0.480	 &0.306	 &0.379  &0.366	 &\textbf{0.698}   \\
\hline
\textbf{AR} (URL) & & / & / & / & / & / & / &
        \rev{/} &  3.33 & 3.56 &  4.71 & 4.06    &  4.33 &
       \textbf{1.00}$^\dag$    \\
\textbf{AR} (All) & & \rev{7.83} & \rev{6.15} & \rev{5.23} & \rev{6.46} & \rev{7.35} & \rev{2.17} & \rev{10.60} & \rev{7.38} & \rev{7.12} & \rev{10.62} & \rev{8.98} & \rev{9.19} &  \rev{\textbf{1.92}$^\dag$} \\
\hline
\makecell[l]{\eat{\textbf{1 vs. 1 Comp.} \\ }\textbf{W/T/L}} & & 22/0/2 & 22/0/2 & 21/0/3 & 21/1/2 & 22/0/2 & 12/3/9 & \rev{24/0/0} & 24/0/0 & 24/0/0 & 24/0/0 & 24/0/0 & 24/0/0 &  / \\
\hline
\textbf{p-val} & &\underline{0.0000} &  \underline{0.0000} & \underline{0.0000} & \underline{0.0002} & \underline{0.0000}  & 0.9169 & \rev{\underline{0.0000}} & \underline{0.0000} & \underline{0.0000} & \underline{0.0000} & \underline{0.0000} & \underline{0.0000} & / \\
    \bottomrule
    \end{tabular}}    
\end{table*}

We reproduce the URL baselines using the open source code from the authors' implementations with the recommended configurations. The results of the classification baselines and the task-specific clustering baselines are taken from the published papers~\cite{UEA,ts2vec,OSCNN,DSN,Shapenet,MUSLA}. Other results are based on our reproduction.

 \noindent
\textbf{Metrics.}
{Standard metrics are employed to evaluate the performance of the downstream tasks. We utilize Accuracy (Acc)~\cite{UEA} in classification tasks. Clustering results are evaluated using Rand Index (RI) and Normalized Mutual Information (NMI)~\cite{MUSLA,USSL}. And F1-score is adopted for anomaly detection~\cite{ASD-KDD21,anomaly_detection_kdd22}.
}

\vspace{-1.5ex}
\subsection{Main Results}

Table~\ref{tab:classification}, \ref{tab:clustering} and \ref{tab:anomaly_detection} summarize the results on classification, clustering, and anomaly detection tasks. We report the average ranking (AR) of algorithms on each dataset, and count the number of datasets in which
the CSL wins/ties/loses (W/T/L) the counterparts in the one-versus-one comparisons. The Wilcoxon rank test’s p-values (p-val) are employed
to quantificationally evaluate the significance.

In summary, the proposed CSL outperforms the URL competitors on most of the tasks and datasets, achieving the best overall performance. Moreover, CSL can achieve performance comparable to the approaches customized for classification and clustering. The results show the excellent ability of CSL in unsupervised learning of high-quality and general-purpose MTS representation.
Below we discuss the results in detail for each task.   

\noindent
\textbf{Classification.}
As shown in Table~\ref{tab:classification}, CSL achieves competitive performance on most of the datasets. It has the highest average accuracy and accuracy ranking. Specifically, among the 30 datasets, CSL achieves the best accuracy in 21 of them if compared to URL methods only, and the highest accuracy in 12 of them (best in all algorithms) if all methods are considered. In the one-versus-one comparison, CSL outperforms all URL competitors in terms of the number of wins on the datasets. These results are in line with our expectations, as shapelets are originally designed to extract time series patterns that can effectively distinguish different classes. CSL further enhances the advantages of shapelets by jointly using the shapelets of different lengths and multiple (dis)similarity measures, and by using a novel objective for model training. To our surprise, CSL achieves better overall accuracy than the fully supervised counterparts. Compared to the supervised learning methods and based on the Wilcoxon rank test, our CSL has surpassed MSLTM-FCNs and TapNet and is on par with ShapeNet, OSCNN, and DSN, showing that CSL has reached a comparable level to supervised learning. This implies that class-specific features can be learned from the inherent structure of the data without supervised information, thus labels are only needed for classifier training.  Furthermore, we observe that CSL performs poorly on DuckDuckGeese (DD), which has a very high dimension of 1345 (see Table~\ref{tab:uea_statistics}). This may indicate a relative weakness of  CSL in dealing with high-dimensional MTS, which is a possible direction to further improve our method.

\noindent
\textbf{Clustering.}
The results of the clustering tasks are shown in Table~\ref{tab:clustering}. CSL outperforms all the other competitors except MUSLA. We note that the best performance for most of the datasets is achieved by either CSL or MUSLA, which are both based on time series shapelet methods. This result shows the superiority of shapelet features for the MTS clustering tasks. CSL outperforms MUSLA in terms of average ranking, the number of best performances, and the number of wins in one-versus-one comparisons, while slightly underperforming MUSLA in terms of average RI and NMI. Overall, there is no statistically significant difference between these two methods. We would like to emphasize that MUSLA is a specialized clustering method, while our CSL is a generic URL algorithm that can be used for a variety of downstream tasks. MUSLA also relies on exhaustive search or prior knowledge to determine the length of shapelets, while CSL can achieve comparable performance without any effort in this regard. \rev{Besides, we notice that SN-C has almost the worst overall performance, which indicates that the URL-based shapelet selection method of ShapeNet which is customized for classification cannot be well generalized to the clustering problem.}

\begin{table}[t]
        \centering
        \caption{Performance comparison on MTS anomaly detection. $w$ represents the length of the sliding window and the best results are highlighted in bold. The underlined value indicates significant difference under a statistical level of 0.05.}
        \vspace{-2.5ex}
        \label{tab:anomaly_detection}
        \resizebox{\linewidth}{!}{
        \begin{tabular}{lcccccccccc}
        \toprule
         \textbf{Dataset} & w & IF-s & IF-p & \rev{SN-AD} & TS2Vec & T-Loss & TNC & TS-TCC & TST & \textbf{CSL} \\
         \midrule
         \multirow{4}{*}{SMAP} & 25 
&0.1040 
&0.2146 
&\rev{0.2408}
&0.2680 
&0.3479 
&0.3111 
&0.3383 
&0.2279 
&\textbf{0.4088}\\
          & 50 
&0.0982 
&0.2090 
&\rev{0.2253}
&0.3096 
&0.3834 
&0.3089 
&0.3662 
&0.2399 
&\textbf{0.3989}\\
          & 75
&0.0377 
&0.2149 
&\rev{0.2195}
&0.2365 
&0.3806 
&0.3133 
&0.3578 
&0.2383 
&\textbf{0.3964}\\
          & 100
&0.0890 
&0.2166 
&\rev{0.2744}
&0.2834 
&0.3862 
&0.3218 
&0.3501 
&0.2511 
&\textbf{0.4049}\\
          \hline
         \multirow{4}{*}{MSL} & 25 
&0.0212 
&0.2235 
&\rev{0.2408}
&0.1557 
&0.2300 
&0.2097 
&0.2420 
&0.2104 
&\textbf{0.3312}\\
          & 50 
&0.0160 
&0.2375 
&\rev{0.2571}
&0.1422 
&0.2563 
&0.2485 
&0.2744 
&0.2292 
&\textbf{0.3725}\\
          & 75
&0.0092 
&0.2522 
&\rev{0.2796}
&0.1588 
&0.2661 
&0.2513 
&0.2898 
&0.2474 
&\textbf{0.3813}\\
          & 100
&0.0077 
&0.2653 
&\rev{0.2907}
&0.1753 
&0.2771 
&0.2629 
&0.3074 
&0.2645 
&\textbf{0.4033}\\
          \hline
         \multirow{4}{*}{SMD} & 25
&0.2453 
&0.1664 
&\rev{0.1807}
&0.2035 
&0.1685 
&0.1754 
&0.1870 
&0.1394 
&\textbf{0.2723}\\
          & 50
&0.2666 
&0.1799 
&\rev{0.1981}
&0.2345 
&0.1955 
&0.1867 
&0.2140 
&0.1416 
&\textbf{0.2777}\\
          & 75 
&0.2715 
&0.1963 
&\rev{0.1953}
&0.2659 
&0.2211 
&0.2070 
&0.2327 
&0.1704 
&\textbf{0.2782}\\
          & 100
&\textbf{0.2912}
&0.2152 
&\rev{0.2101}
&0.2846 
&0.2472 
&0.2291 
&0.2454 
&0.1859 
&0.2784\\
          \hline
         \multirow{4}{*}{ASD} & 25
&0.1916 
&0.1878 
&\rev{0.1403}
&0.2978 
&0.3223 
&0.1591 
&0.1938 
&0.1351 
&\textbf{0.3272}\\
          & 50
&0.2149 
&0.2368 
&\rev{0.2084}
&0.3412 
&0.3504 
&0.1812 
&0.2324 
&0.1933 
&\textbf{0.3799}\\
          & 75 
&0.2391 
&0.2782 
&\rev{0.2488}
&0.3655 
&0.3544 
&0.2367 
&0.2844 
&0.2398 
&\textbf{0.4094}\\
          & 100
&0.2585 
&0.3193 
&\rev{0.3062}
&0.3735 
&0.3449 
&0.2945 
&0.2679 
&0.2926 
&\textbf{0.4349}\\
\hline
\textbf{Avg F1}  
&
&0.1476 
&0.2258 
&\rev{0.2323} 
&0.2560 
&0.2957 
&0.2436 
&0.2740 
&0.2129 
&\textbf{0.3597}\\ 

\hline
\textbf{AR} & & \rev{6.69} & \rev{6.38} & \rev{5.63} & \rev{4.56} & \rev{3.50} &  \rev{6.13} & \rev{3.63} & \rev{7.38} &
       \rev{\textbf{1.13}}\\

\hline
\makecell[l]{\eat{\textbf{1 vs. 1 Compasion} \\} \textbf{W/T/L}} & & 15/0/1 & 16/0/0 & \rev{16/0/0} & 15/0/1 & 16/0/0 & 16/0/0 & 16/0/0 & 16/0/0 &  / \\

\hline
\textbf{p-val} & & \underline{0.0002} & \underline{0.0000} & \rev{\underline{0.0000}} & \underline{0.0001} & \underline{0.0000} & \underline{0.0000} & \underline{0.0000} & \underline{0.0000} & /\\
          \bottomrule
        \end{tabular}
        }
        \vspace{-3ex}
\end{table}

\noindent
\textbf{Anomaly Detection.}
In Table~\ref{tab:anomaly_detection}, we can see that CSL outperforms the baselines in every setting, except for SMD with a window length of 100, where CSL is slightly inferior to IF-s and TS2Vec. This may indicate that these two algorithms are more effective in detecting outliers with long SMD windows. For each dataset, the performance of all methods tend to improve as the sliding window size increases, because larger windows allow more normal observations to be seen to better detect the outliers. CSL achieves superior performance on the MSL dataset, outperforming the second-best TS-TCC by more than 30\% on each window size. Although the difference in performance is not as large on the other three datasets, CSL is almost always the best and significantly outperformed each competitor. This indicates that CSL has an outstanding ability to identify anomalies. We also observe that the second-best method is different for each dataset, with T-Loss on SMAP, TS-TCC on MSL, and IF-s on SMD, while TS2Vec and T-Loss are the second-best methods with about the same performance on ASD. The reason may be that these URL algorithms are developed based on specific assumptions that may fail in other domains. In contrast, CSL exhibits more general capabilities. \rev{The variant of ShapeNet, i.e., SN-AD, does not achieve competitive performance like in classification, showing again the limitation of ShapeNet in terms of task generality.}

\rev{Finally, we visualize the unsupervised learned representation of ERing test data using t-SNE~\cite{tSNE}. We compare CSL with the five URL baselines and the variant of ShapeNet which excludes the supervised feature selection. As Fig.~\ref{fig:repr} shows, the representation learned by the proposed CSL forms more separated clusters, which also suggests representation of lower entropy. This explains why CSL can outperform the competitors on downstream analysis tasks.}

\vspace{-2ex}
\subsection{Ablation Study}\label{sec:exp:ablation}  
To validate the effectiveness of the key components in CSL, we conduct ablation studies using classification tasks on all 30 UEA datasets. Due to space limitations, only the statistical results are reported here. \eat{The full results can be found in Appendix~\ref{app:ablation}.} The best {value in a comparison} is highlighted in bold and underlining indicates a significant difference under a statistical level of 0.05. The results are discussed as follows.

\noindent
\textbf{Effectiveness of components in Shapelet Transformer.} There are two major designs within the Shapelet Transformer, including using the shapelets of different scales (lengths) and the diverse dis(similarity) measures. As they improve the representation in orthogonal directions, we assess their effectiveness individually. 

\begin{table}[t]
    \centering
    \caption{Effectiveness of multi-scale shapelets.}
    \vspace{-2.5ex}
    \resizebox{.9\linewidth}{!}{\begin{tabular}{lcccc}
    \toprule
    \textbf{Statistic}  & Short scale only & Long scale only & Better scale & \textbf{Both (CSL)} \\
    \midrule
       Avg Acc/AR  & 0.710/2.88 & 0.687/2.98 & 0.723/2.12  & \textbf{0.735}/\textbf{2.02}\\
    \hline
       \makecell[l]{\eat{\text{1 vs. 1 Compasion} \\ (Wins/Ties/Loses)}W/T/L} & 19/4/7 & 21/3/6 & 13/6/11 & / \\
    \hline
    \text{p-val} & \underline{0.0054} & \underline{0.0001} & 0.1791 & / \\
    \bottomrule
    \end{tabular}}
    \label{tab:ablation_shapelet_scales}
    \vspace{-1ex}
\end{table}

\begin{table}[t]
    \centering
    \caption{Effectiveness of diverse (dis)similarity measures.}
    \vspace{-2.5ex}
    \resizebox{.83\linewidth}{!}{\begin{tabular}{lcccc}
    \toprule
      \textbf{Statistic}   & Euclidean only & Cosine only & Cross only & \textbf{All (CSL)} \\
    \midrule
       Avg Acc/AR  & 0.652/3.33	& 0.698/2.93 &	0.709/2.05 & \textbf{0.735}/\textbf{1.68}\\
    \hline
       \makecell[l]{\eat{\text{1 vs. 1 Compasion} \\ (Wins/Ties/Loses)}W/T/L} & 25/4/1 & 24/1/5 & 17/2/11 & /\\
     \hline
    \text{p-val} &  \underline{0.0000} & \underline{0.0001} & \underline{0.0426} & /\\
    \bottomrule
    \end{tabular}}
    \label{tab:ablation_shapelet_measures}
    \vspace{-1ex}
\end{table}

\begin{table}[t]
    \centering
    \caption{Effectiveness of multi-grained contrasting and multi-scale alignment.}
    \vspace{-2.5ex}
    \resizebox{.73\linewidth}{!}{\begin{tabular}{lcccc}
    \toprule
     \textbf{Statistic}    & w/o  $\mathcal{L}_C$ & w/o  $\mathcal{L}_F$ & w/o  $\mathcal{L}_A$ & \textbf{All (CSL)} \\
    \midrule
       Avg Acc/AR  &  0.720/2.38 & 	0.708/3.33 & 0.709/2.92		& \textbf{0.735}/\textbf{1.37} \\
    \hline
       \makecell[l]{\eat{\text{1 vs. 1 Compasion} \\ (Wins/Ties/Loses)}W/T/L} & 20/5/5 & 27/3/0 & 26/4/0 & /\\
       \hline
    \text{p-val} & \underline{0.0006} & \underline{0.0000} & \underline{0.0000} & /\\
    \bottomrule
    \end{tabular}}
    \label{tab:ablation_losses}
    \vspace{-2ex}
\end{table}

\textit{(1) \underline{Multi-scale shapelets.}} ST contains shapelets ranging from short to long. Here we compare CSL with its three variants: \textit{short scale only} (where the shapelet length ranges from $0.1 T$ to $0.4 T$), \textit{long scale only} (from $0.5 T$ to $0.8 T$) and the \textit{better of the two}. To make a fair comparison, we fix the embedding dimension $D_{repr}$ and the number of scales $R$ for all experiments. The results are shown in Table~\ref{tab:ablation_shapelet_scales}. Both the short- and long-scale variants perform much worse than CSL. Even the best of the two variants still performs slightly worse than CSL. These results demonstrate the necessity of using shapelets with a wider range of time scales.

\textit{(2) \underline{Diverse (dis)similarity measures.}} {
To investigate the role of the dis(similarity) measures in the Shapelet Transformer, we compare our CSL with its three variants, i.e., separately using one measure of the \textit{Euclidean norm}, \textit{cosine similarity}, and \textit{cross correlation}. The results are summarized in Table~\ref{tab:ablation_shapelet_measures}. We can see that the cross correlation is the best performer among the three single measures, while the Euclidean norm variant using the original definition of shapelet is the worst. This validates our hypothesis that the Euclidean norm-based shapelet has limitations in representing time series. All three variants perform much worse than CSL with the p-values less than 0.05. This shows the need to combine the different types of measures in the shapelet-based MTS representation.}


{
\noindent
\textbf{Effectiveness of components in loss function.}
There are three terms in our loss function, i.e., \textit{coarse-grained contrastive loss} $\mathcal{L}_C$, \textit{fine-grained contrastive loss} $\mathcal{L}_F = \sum_{r=1}^R \mathcal{L}_{F, r}$, and \textit{multi-scale alignment loss} $\mathcal{L}_A$. We investigate the effect size of each term by removing them one by one. As we can see in Table~\ref{tab:ablation_losses}, CSL is significantly better than the variant without the term $\mathcal{L}_F$ or $\mathcal{L}_A$, which proves the importance of these two components. In contrast, removing the coarse-grained loss $\mathcal{L}_C$ has the least impact. This may imply that, when the representations on each time scale have been sufficiently trained and aligned, the joint version is already near-optimal, thus the coarse-grained contrasting can no longer lead to a huge (but still statistically significant) improvement like the other two terms. }

\begin{figure}[t]
    \centering
    \includegraphics[width=\linewidth]{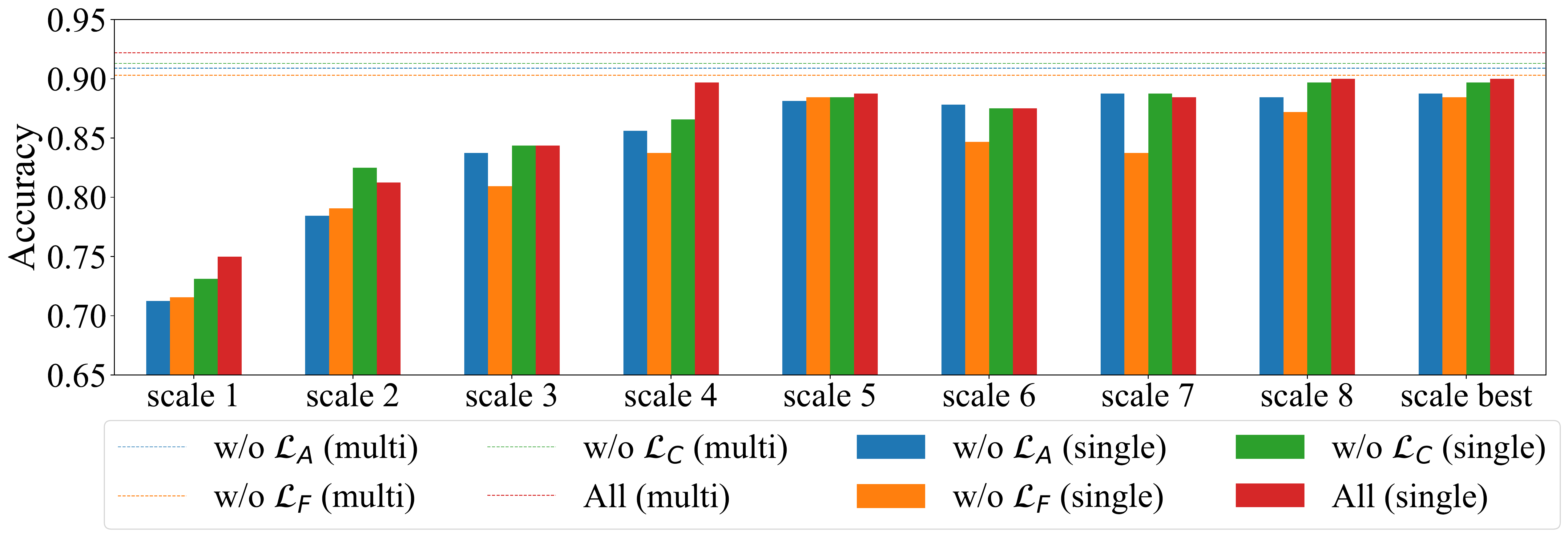}
    \vspace{-5ex}
    \caption{Study of the multi-grained contrasting and the multi-scale alignment on UWaveGestureLibrary. Dashed line corresponds to the joint embedding in $\mathbb{R}^{D_{repr}}$ with multiple scales and bar corresponds to embedding at each single scale.}
    \vspace{-1.5ex}
    \label{fig:learning_objective_UW}
\end{figure}

\begin{table}[t]
    \centering
    \caption{Effectiveness of the data augmentation library.}
    \vspace{-3ex}
    \resizebox{\linewidth}{!}{\begin{tabular}{lcccccc}
    \toprule
    \textbf{Statistic}     & w/o $J(\xv)$ & w/o $C(\xv)$ & w/o $TW(\xv)$ & w/o $Q(\xv)$ & w/o $P(\xv)$ & \textbf{All (CSL)} \\
    \midrule
       Avg Acc/AR  & 0.715/3.63	& 0.716/3.70 &	0.707/4.52 &	0.712/3.88	& 0.718/3.28 &	\textbf{0.735}/\textbf{1.98}\\
    \hline
       \makecell[l]{\eat{\text{1 vs. 1 Compasion} \\ (Wins/Ties/Loses)}W/T/L} & 19/7/4 & 21/7/2 & 24/3/3 & 23/5/2 & 19/7/4 & / \\
       \hline
       \text{p-val} & \underline{0.0004} & \underline{0.0003} & \underline{0.0001} & \underline{0.0001} & \underline{0.0006} & / \\
    \bottomrule
    \end{tabular}}
    \label{tab:ablation_augmentations}
    \vspace{-3ex}
\end{table}


We further explore how multi-grained contrasting and multi-scale alignment work using a case study in Fig.~\ref{fig:learning_objective_UW}. \eat{We find that CSL outperforms the variants without $\mathcal{L}_F$ or $\mathcal{L}_A$ on most single scales, indicating the importance of fine-grained contrasting and multi-scale alignment across different scales.} \zy{We find that removing $\mathcal{L}_F$ decreases the representation quality of every single scale (the orange bar), and thus has a great negative impact on the joint embedding (the orange line). Similar phenomena can be observed for $\mathcal{L}_A$. It indicates that $\mathcal{L}_F$ and $\mathcal{L}_A$ improve the final performance through improving the quality of each scale. Compared to the variants without $\mathcal{L}_C$  (the green bar), the representation quality of each single scale is \textit{balanced} with the loss (the red bars), saying that the quality of scale 1, 4, 5 and 8 is improved, while the quality of scale 2 and 7 is a little decreased. As a result, the joint embedding learned using $\mathcal{L}_C$  (the red line) is better than that without the loss (the green line). It validates the hypothesis in Section~\ref{sec:multi-grained-contrasting}  that $\mathcal{L}_C$  can coordinate the multiple scales to improve the joint embedding.}    \eat{Removing $\mathcal{L}_C$ reduces accuracy at scales 1, 4, 5, and 8, but has no negative impact on performance at the other four scales. However, without $\mathcal{L}_C$, the performance of the joint embedding is much lower. This suggests that coarse-grained comparisons still play a role in the joint embedding space. Note that the performance of the best single scale is still worse than the performance of using all scales jointly. This is the reason why CSL explores the representation of multiple scales.
}

{From the above exhaustive analysis, we can conclude that all the components included in the loss function of CSL are necessary.}

\noindent
\textbf{Effectiveness of the data augmentation library.}
{
We remove the methods in the data augmentation library one by one to evaluate their effectiveness. As can be seen in Table~\ref{tab:ablation_augmentations}, the variant without time warping get the lowest average ranking (4.52), implying that removing time warping has a broader negative effect among the 30 datasets than the other data augmentation methods. The data augmentation libraries without each of the other four methods have a close average ranking\eat{because they have different effects on different datasets (see Table~\ref{tab:ablation_augmentations_full} in Appendix)}. In contrast, the complete version has the highest average ranking (1.98), suggesting that the performance of CSL can probably be further improved when more types of data augmentation approaches are included in the library. This is an interesting finding and may imply a general data-independent data augmentation scheme for unsupervised representation learning of MTS. We leave the further exploration in our future work.
}

\vspace{-0.5ex}
\zy{\subsection{Sensitivity Analysis}}
We perform sensitivity analysis to study the key parameters, including the number of shapelet scales $R$ (default 8), the  minimum and maximum lengths of the shapelets $L_{min}$ (default $0.1T$) and $L_{max}$ (default $0.8T$), the decay rate $\alpha$ (default 0.5), and the regularization coefficients $\lambda$ (default 0.01) and $\lambda_S$ (default 1). 

\begin{figure}[t]
    \centering
\subcaptionbox{\centering \# Scales $R$.
    \label{subfig:R}}
{\includegraphics[width=.32\linewidth]{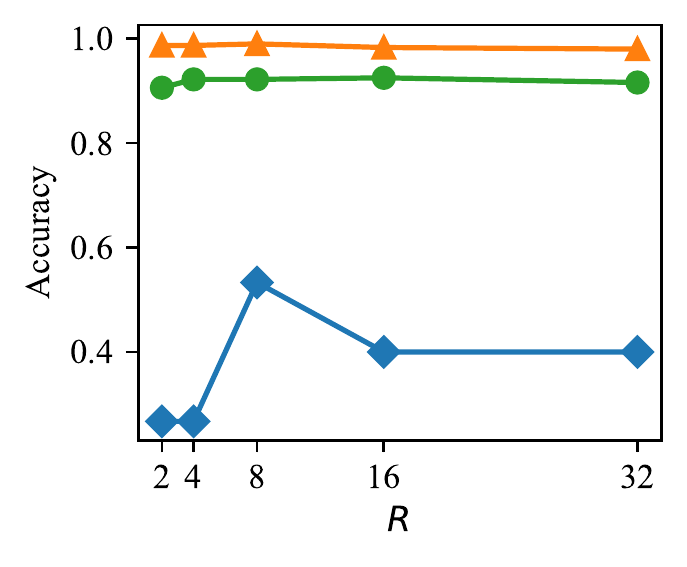}\vspace{-1.2ex}}
\subcaptionbox{\centering Min. length $L_{min}$.
    \label{subfig:Lmin}}
{\includegraphics[width=.32\linewidth]{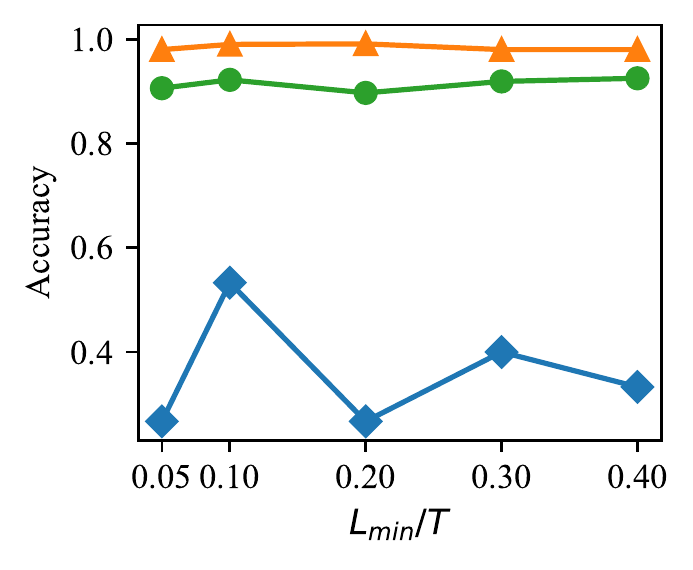}\vspace{-1.2ex}}
\subcaptionbox{\centering Max. length $L_{max}$.
    \label{subfig:Lmax}}
{\includegraphics[width=.32\linewidth]{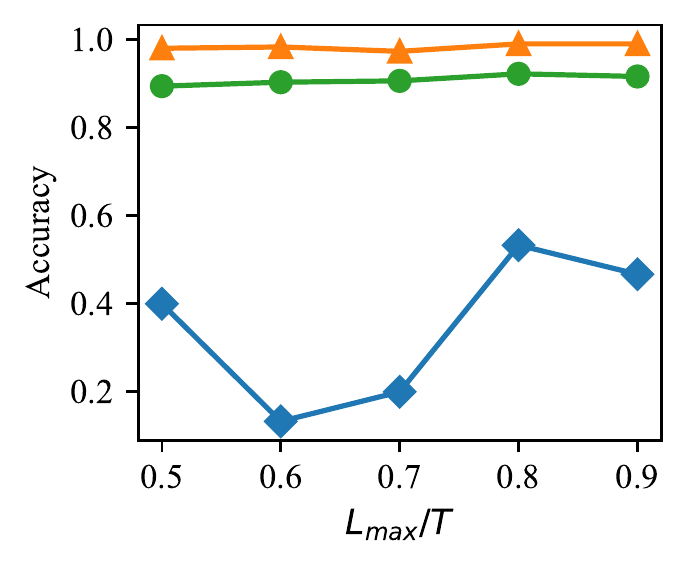}\vspace{-1.2ex}}
\\ \vspace{0.5ex}
\subcaptionbox{\centering Decay rate $\alpha$.
    \label{subfig:alpha}}
{\includegraphics[width=.32\linewidth]{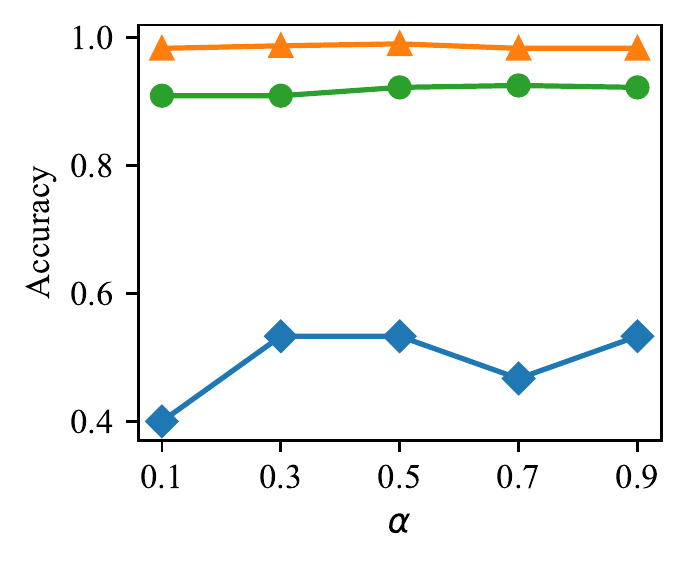}\vspace{-1.2ex}}
\subcaptionbox{\centering  $\mathcal{L}_A$ importance  $\lambda$.
    \label{subfig:lmd}}
{\includegraphics[width=.32\linewidth]{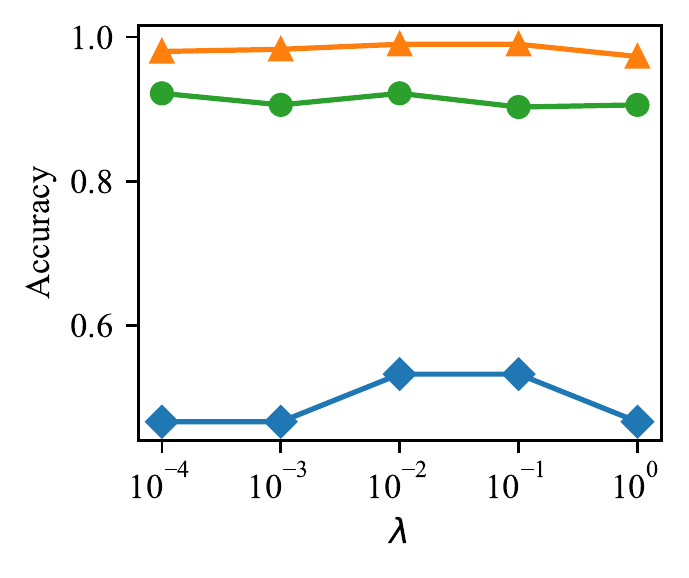}\vspace{-1.2ex}}
\subcaptionbox{\centering $\mathcal{L}_S$ importance $\lambda_S$.
    \label{subfig:lmds}}
{\includegraphics[width=.32\linewidth]{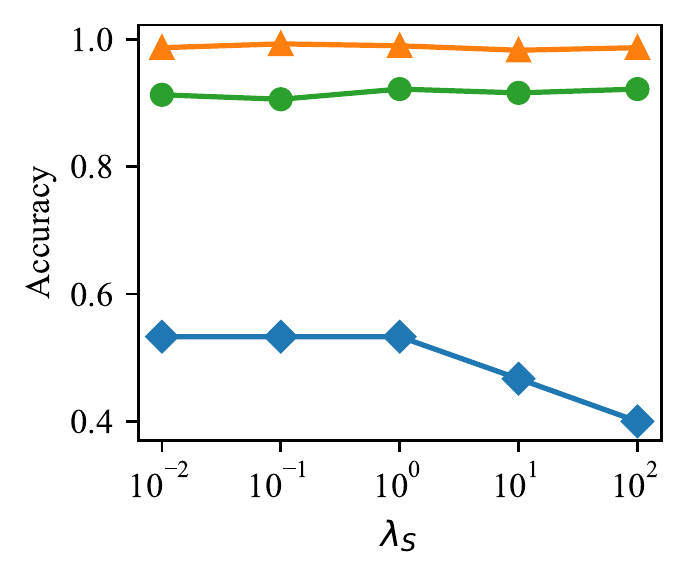}\vspace{-1.2ex}}
\captionsetup[subfigure]{labelformat=empty}
\\ \vspace{-3ex}
\subcaptionbox{
    \label{subfig:legend}}\centering
{\includegraphics[width=.93\linewidth]{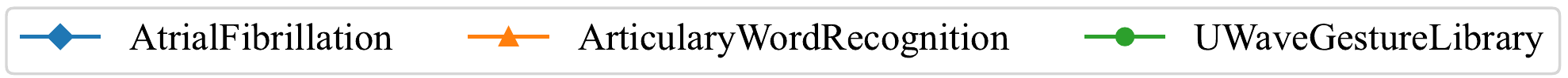}}
\vspace{-6.5ex}
    \caption{Sensitivity analysis of the key parameters.}
    \label{fig:sen_analysis}
    \vspace{-2ex}
\end{figure}

Similar to the setting in Section 3.2, given $L_{min}$, $L_{max}$ and $R$, the shapelet lengths are simply set to the evenly spaced numbers over $[L_{min}, L_{max}]$, i.e., $L_r = L_{min} + (r - 1)\frac{L_{max} - L_{min}}{R-1}$ ( $r \in \{1,\ldots,R\}$). 
The performance is evaluated using classification accuracy. The results on three diverse UEA datasets are shown in Fig.~\ref{fig:sen_analysis} and the similar trends can be observed on the other datasets. Please note that the performance on AtrialFibrillation seems sensitive just because the dataset has only 15 testing samples, the minimum number among the 30 UEA datasets, which is a corner case of our evaluation. We discuss the results in detail as follows. 

\noindent
\textbf{The sensitivity analysis of $R$.}
To capture multi-scale information, the number of scales $R$ cannot be too small. But too large $R$ will cause a small number of shapelets $K$ under a fixed representation dimension $D_{repr}$, which can also decrease the representation quality. As the result in Fig.~\ref{subfig:R} shows, the model is relatively more sensitive to small values of $R$ than larger values, while a moderate value of $8$ can lead to good overall performance among the datasets. 

\noindent
\textbf{The sensitivity analysis of $L_{min}$ and $L_{max}$.}
From Fig.~\ref{subfig:Lmin}-\ref{subfig:Lmax}, we can see that for UWaveGestureLibrary, the best choice of $L_{min}$ is about $0.4T$ and the values of $0.8T$-$0.9T$ are the best for $L_{max}$, which indicates that the long-term features can be more effective than the short-term ones. For ArticularyWordRecognition, the relatively small values of $L_{min}$ ($0.1T$-$0.2T$) and large values of $L_{max}$ ($0.8T$-$0.9T$) are better, showing the importance of both short- and long-term features. While on the AtrialFibrillation dataset, $L_{min} = 0.1T$ and $0.8 T$-$0.9 T$ for $L_{max}$ are empirically the best choice. Overall, the default settings of $L_{min} = 0.1T$ and $L_{max} = 0.8T$  can be decent for different datasets without any tunning (also validated in Section~\ref{sec:exp:ablation}), while one can manually optimize them for further improvement.

\begin{figure}
    \centering
    \includegraphics[width=0.47
    \linewidth]{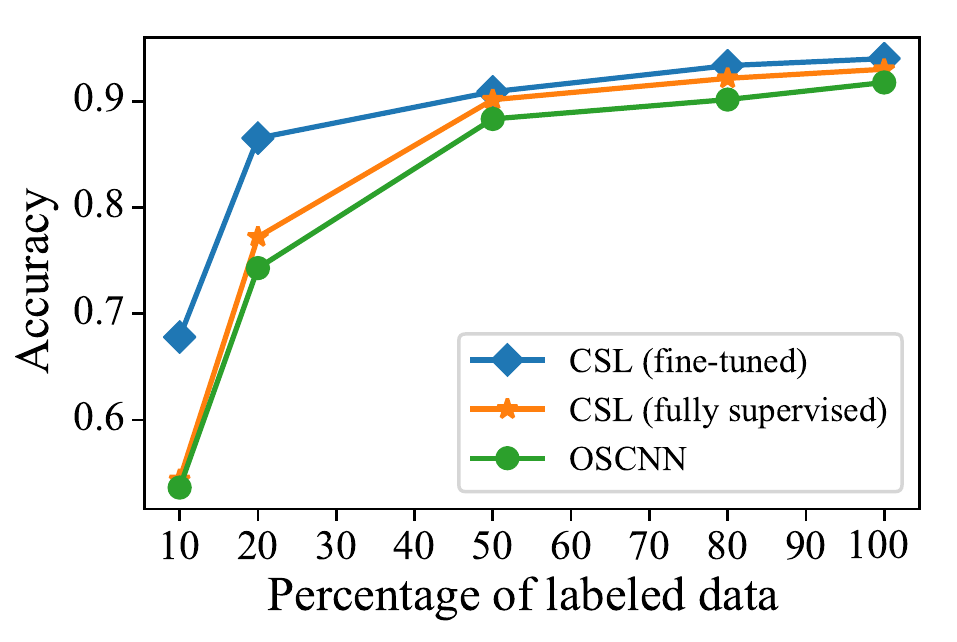}
    \vspace{-2ex}
    \caption{Accuracy of OSCNN, fully supervised and fine-tuned CSL w.r.t. the ratio of labeled data on UWaveGestureLibrary.}
    \label{fig:sparse_labeled_UW}
    \vspace{-3ex}
\end{figure}

\noindent
\textbf{The sensitivity analysis of $\alpha$.}
As the result shown in Fig.~\ref{subfig:alpha}, our model is more sensitive to the small values of the decay rate $\alpha$ than the large values for the UWaveGestureLibrary and AtrialFibrillation datasets, and the opposite for ArticularyWordRecognition. Overall, our model is less sensitive to $\alpha$ than the other parameters, and a moderate value around $0.5$ is better for different datasets. 

\noindent
\textbf{The sensitivity analysis of $\lambda$ and $\lambda_S$.}
As shown in Fig.~\ref{subfig:lmd}, by varying $\lambda$ from 1 to 0.0001, we observe that our model achieves good performance among different datasets when $\lambda$ is around $0.01$. Similarly, we vary $\lambda_S$ from 100 to 0.01. The result in Fig.~\ref{subfig:lmds} indicates that our model is more robust to the larger values of $\lambda_S$ (1 to 100) for UWaveGestureLibrary and the opposite for AtrialFibrillation (where the model is more sensitive when $\lambda_S > 1$). Overall, a moderate value around 1 can be a good choice for different datasets.

\begin{figure}[t]
    \centering
    \subcaptionbox{Time series of the four classes and two shapelets with different lengths and (dis)similarity measures learned by CSL.
    \label{subfig:shapelet-matching}}
{\includegraphics[width=.48\linewidth]{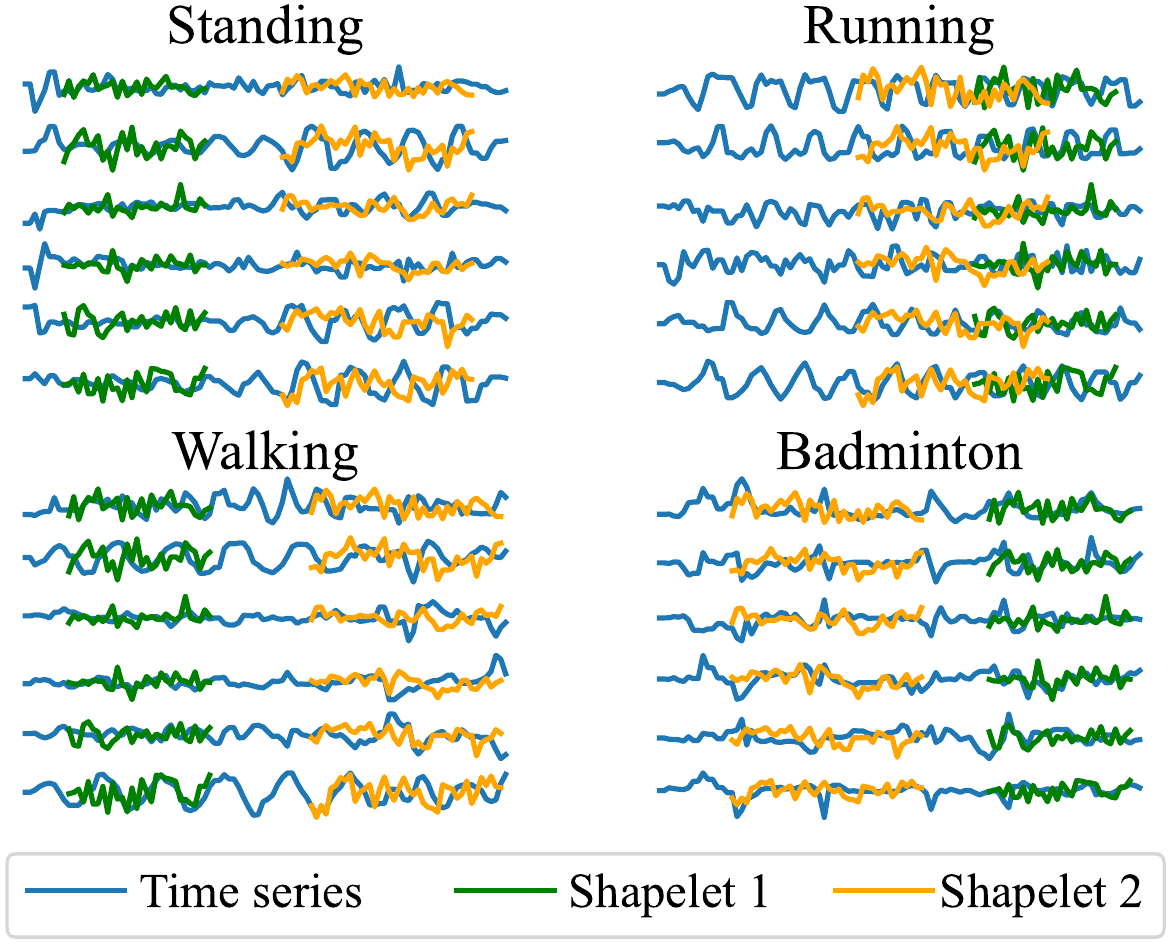}\vspace{-1ex}}
\hspace{1ex}
    \subcaptionbox{The two-dimensional representations of all time series encoded using the two learned shapelets. \label{subfig:shapelet-transformation}\vspace{-1ex}}
{\includegraphics[width=.38\linewidth]{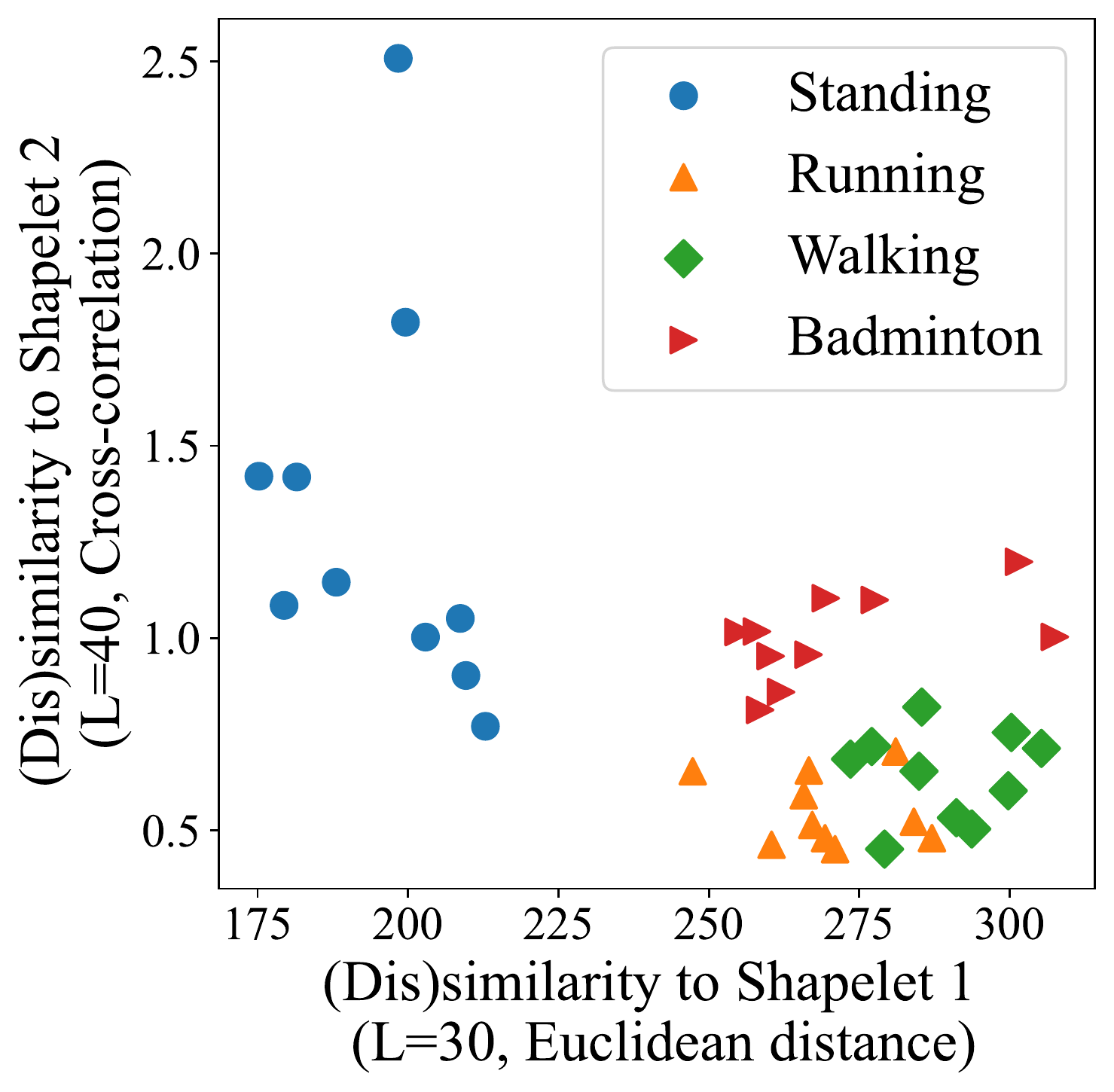}}
\vspace{-3ex}
    \caption{Explanation of the shapelets learned by CSL.}
    \label{fig:explanation}
    \vspace{-3ex}
\end{figure}

\vspace{-0.8ex}
\subsection{Study of Partially Labeled Classification}

To further demonstrate the superiority of our CSL, we perform a case study on UWaveGestureLibrary under a practical setting of partially labeled MTS classification. Specifically, we compare CSL with the best-performing supervised OSCNN on the dataset where only a portion of the randomly selected data is labeled. For CSL, we first use all the data to train the Shapelet Transformer without using labels. Then, we append a linear classifier on top of the representations, and fine-tune the encoder and linear layer using the available labeled data by minimizing the standard cross-entropy loss as used in OSCNN. In contrast, OSCNN is supervisedly trained using the same labeled data (fully supervised). For comparison, we also train a CSL model in the same fully supervised way as OSCNN. 

As Fig.~\ref{fig:sparse_labeled_UW} shows, the fully supervised CSL performs very closely to OSCNN. The fine-tuned CSL consistently outperforms the two competitors, especially when the proportion of labeled data is small.  Taking advantage of URL which can ``pre-train'' the encoder using all available data regardless of annotations, the fine-tuned CSL uses only 20\% labeled data to achieve accuracy comparable to the fully supervised OSCNN and CSL trained with 50\% labeled data. The results show the superiority of our URL method in partially-label settings,  compared to the traditional fully supervised techniques.

\vspace{-0.8ex}
\zy{\subsection{Study of the Learned Shapelets}}
To provide an intuitive understanding of the features CSL extracts, we study the learned shapelets using an easy-to-understand BasicMotions problem from UEA archive. The time series are sensor records of four human motions, i.e., \textit{Standing}, \textit{Walking}, \textit{Running} and \textit{Badminton}. Each sample has six dimensions and of length $T=100$.

We plot four time series of the four classes and two shapelets with different lengths and measures learned by our CSL (see Fig.~\ref{subfig:shapelet-matching}). Shapelet 1 is of length 30 and encodes the samples using Euclidean distance (Green). Shapelet 2 has a length of 40 and is used in conjunction with the cross-correlation function for encoding (Orange). The shapelets are matched to the most similar subsequences for each of the time series samples.    Note that CSL uses multivariate shapelets to jointly capture the information among different variables, where each shapelet has the same dimensions as the time series. The two shapelets encode each time series sample into a two-dimensional representation (Fig.~\ref{subfig:shapelet-transformation}), where each axis is the (dis)similarity between the shapelet and the matching subsequence according to Eq.~\eqref{eq:general_shapelet}. 

From Fig.~\ref{fig:explanation}, we observe that the representation based on Shapelet 1 (X-axis) can distinguish the \textit{Standing} motion while the other three motions can be effectively classified by the features extracted using both shapelets. \eat{Thus, the shapelets can be seen as the prototypes of some classes, and the representations are explained as the degree the shapelets exist in the time series.} Thus, the shapelets can be seen as the prototypes of some classes and the representations are explained as the degree the shapelets exist in the time series, which is intuitive to understand. Our proposal not only extends the original shapelet which is designed only for supervised classification to general-purpose URL, but also retains its benefit in terms of explainability or interpretability~\cite{ye2011time}. \rev{Although the interpretation method is ad-hoc, it remains a nice property of the shapelet compared to the complex neural networks which are harder to explain~\cite{Interpretable-ML-Book}.}

\begin{table}[t]
    \centering
    \caption{\rev{Accuracy and SVM training time on long time series.}}
    \vspace{-2.5ex}
    \resizebox{\linewidth}{!}{\begin{tabular}{lc|c|cccccc|cc}
    \toprule
      \multirow{2}{*}{\textbf{Dataset}}  & \multicolumn{2}{c|}{\textbf{Raw values}} & \multicolumn{7}{c}{\textbf{Unsupervised learned representation}} \\
       \cline{2-10} \rule{0pt}{2.5ex} & Acc & \textit{Time} & TS2Vec & T-Loss & TNC & TS-TCC & TST & \textbf{CSL} & \textit{\textbf{CSL} (Speedup)} \\
    \midrule
BH 	&\textbf{0.732} & \textit{0.6204s} &  0.712& 	\textbf{0.732}& 	\textbf{0.732}& 	0.658& 	0.720& 	\textbf{0.732} & \textit{\textbf{0.0038s} (163\textnormal{x})}\\ 
CD	        &0.482 & \textit{0.4369s} &  0.520& 	0.695& 	0.549& 	0.629& 	0.625& 	\textbf{0.712} & \textit{\textbf{0.0031s} (141\textnormal{x})}\\ 
DA       &0.240 & \textit{0.1968s} &  0.193&    \textbf{0.400}&  0.220&  0.286&  0.307&  0.373 & \textit{\textbf{0.0012s} (164\textnormal{x})}\\ 
US	        &0.210 & \textit{205.7195s} &  0.155& 	0.600& 	0.112& 	0.379& 	0.183& 	\textbf{0.692} & \textit{\textbf{1.2834s} (160\textnormal{x})}\\ 
    \bottomrule
    \end{tabular}}
    \label{tab:long_series}
    \vspace{-3ex}
\end{table}

\eat{From the above description, we can see that the shapelets learned by CSL are meaningful for representing the time series and intuitive to understand. Our proposal not only extends the original shapelet which is designed only for supervised classification to general-purpose URL, but also retains its benefit in terms of explainability/interpretability~\cite{ye2011time}. \rev{Although the interpretation method is ad-hoc, it remains a nice property of the shapelet compared to the complex neural networks which are harder to explain~\cite{Interpretable-ML-Book}.}}

    

\vspace{-0.8ex}
\subsection{\rev{Study of Long Time Series Representation}}

\rev{We assess the ability of the URL methods on long time series representation. Four datasets from the Time Series Machine Learning Website~\cite{TSC-web} are used including BinaryHeartbeat (BH), CatsDogs (CD), DucksAndGeese (DA) and UrbanSound (US). The series lengths of the four datasets are 18530, 14773, 236784 and 44100 respectively and the other statistics can be found on the website. Following Section 5.1, we train an SVM using either the unsupervised learned representation or the raw values of the training data, and report the test accuracy and the SVM training time (marked in italics) in Table~\ref{tab:long_series}. In terms of accuracy, CSL performs the best on three data sets and the second-best on DA, showing its higher ability in long series representation. T-Loss is also well-performed, but is still inferior to CSL on CD and US. TS2Vec and TST cannot handle long series due to high space complexity, so they have to shorten the raw data by truncation or subsampling following~\cite{ts2vec}, which may cause information loss and result in their low performance. Compared to analysis on the raw values, using the representation learned by CSL can not only improve the accuracy, but also achieve more than 140x of speedups for the SVM training. This indicates the superiority of the proposed CSL in long time series analysis.}

\vspace{-0.8ex}
\subsection{\rev{Running Time Analysis}}\label{sec:exp:time}
\begin{figure}
    \centering
\subcaptionbox{\centering HandMovementDirection.
    \label{subfig:HandMovementDirection}}
{\includegraphics[width=.43\linewidth]{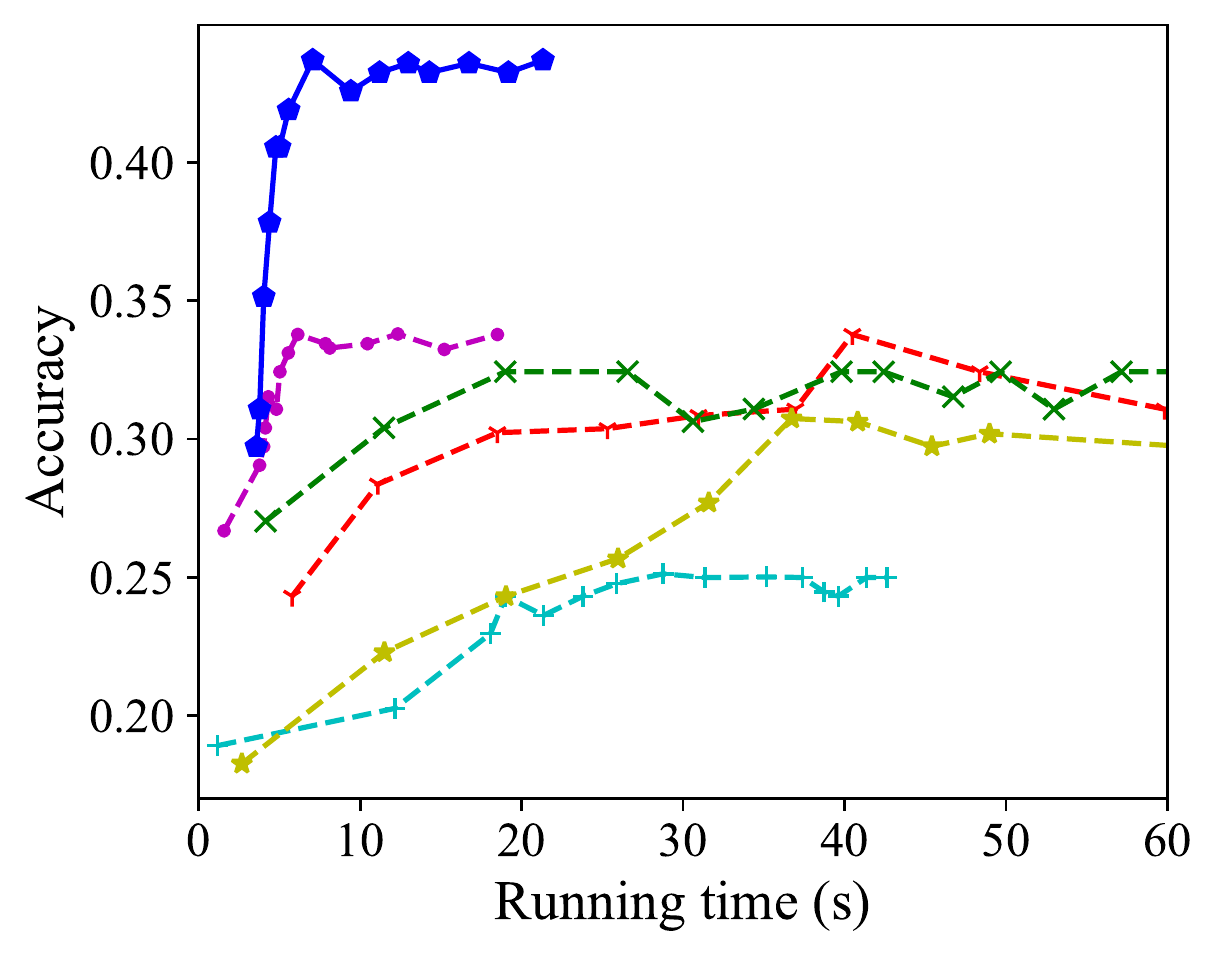}\vspace{-1.5ex}}
\subcaptionbox{\centering Handwriting.
    \label{subfig:Handwriting}}
{\includegraphics[width=.42\linewidth]{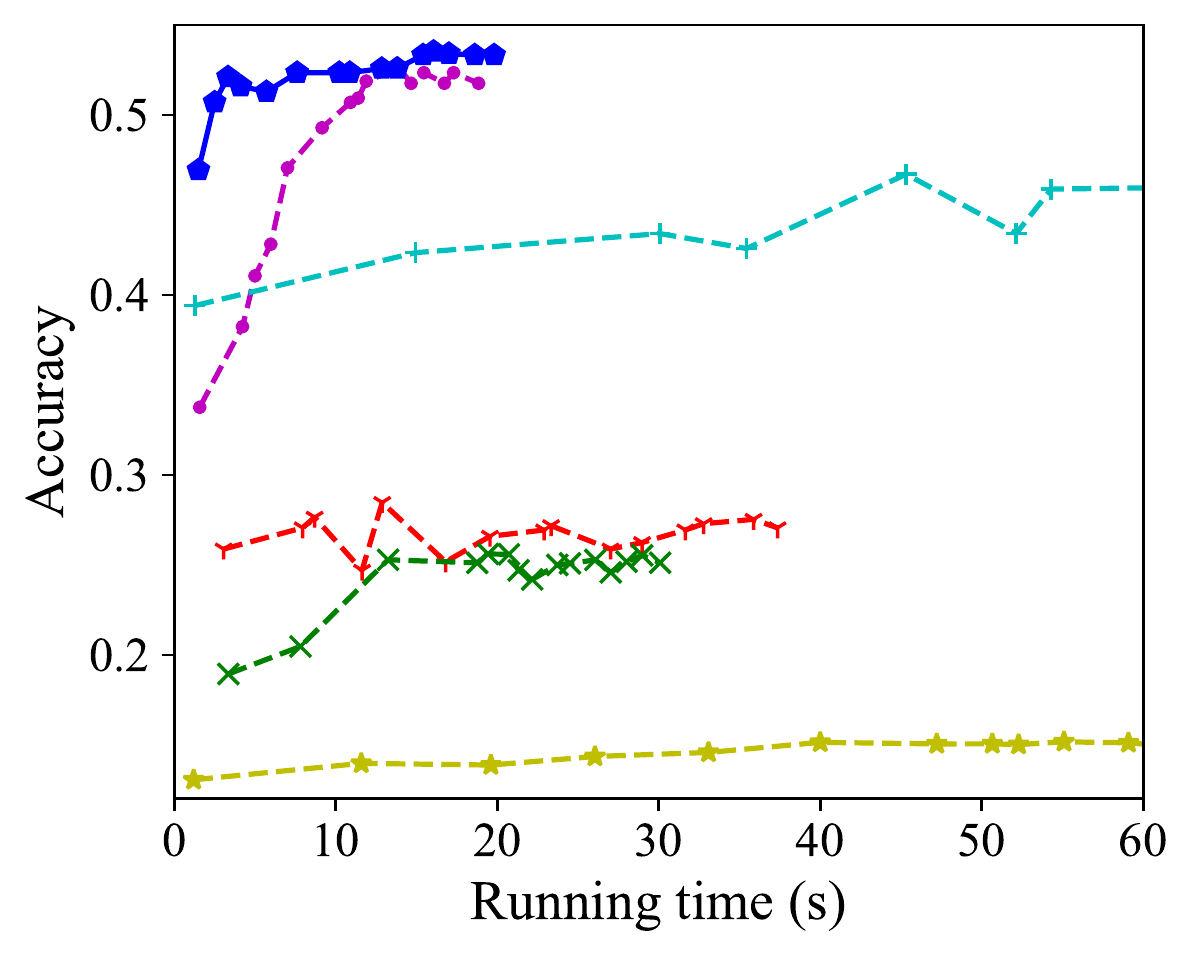}\vspace{-1.5ex}}
\captionsetup[subfigure]{labelformat=empty}
\\ \vspace{-3ex}
\subcaptionbox{
    \label{subfig:legend_acc_time}}\centering
{\includegraphics[width=.8\linewidth]{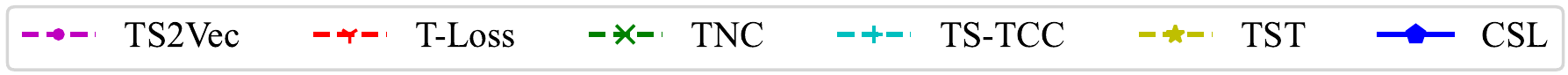}}
\vspace{-6.75ex}
    \caption{\rev{Accuracy w.r.t. total training time.}}
    \label{fig:acc-time}
    \vspace{-2ex}
\end{figure}

\begin{figure}
    \centering
\subcaptionbox{\centering Time w.r.t. $N$.
    \label{subfig:time-N}}
{\includegraphics[width=.325\linewidth]{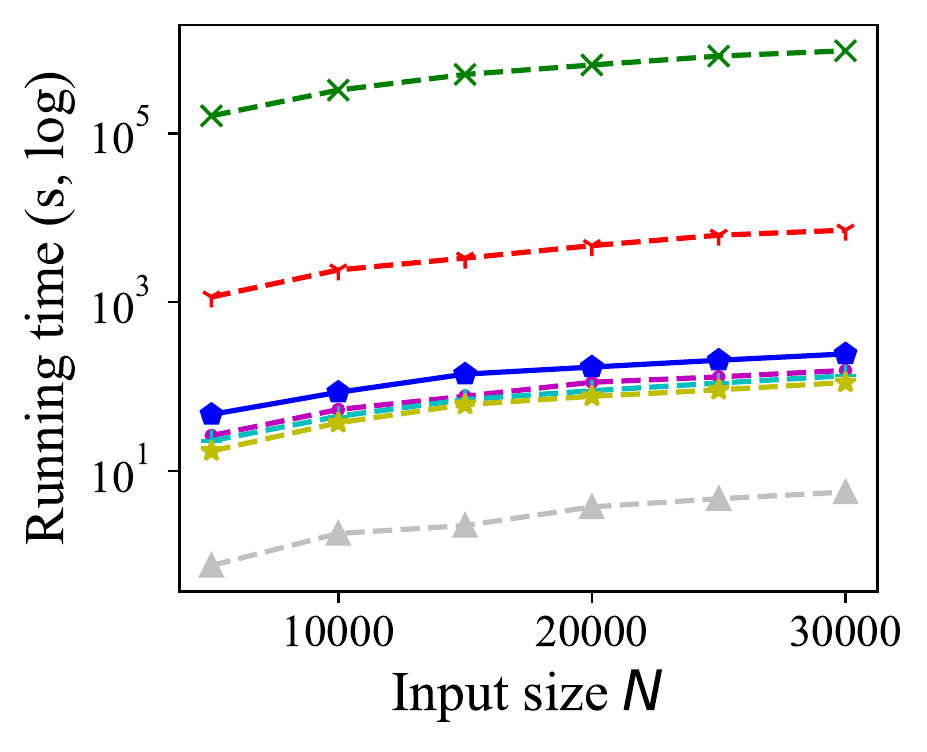}\vspace{-1.75ex}}
\subcaptionbox{\centering Time w.r.t. $D$.
    \label{subfig:time-D}}
{\includegraphics[width=.33\linewidth]{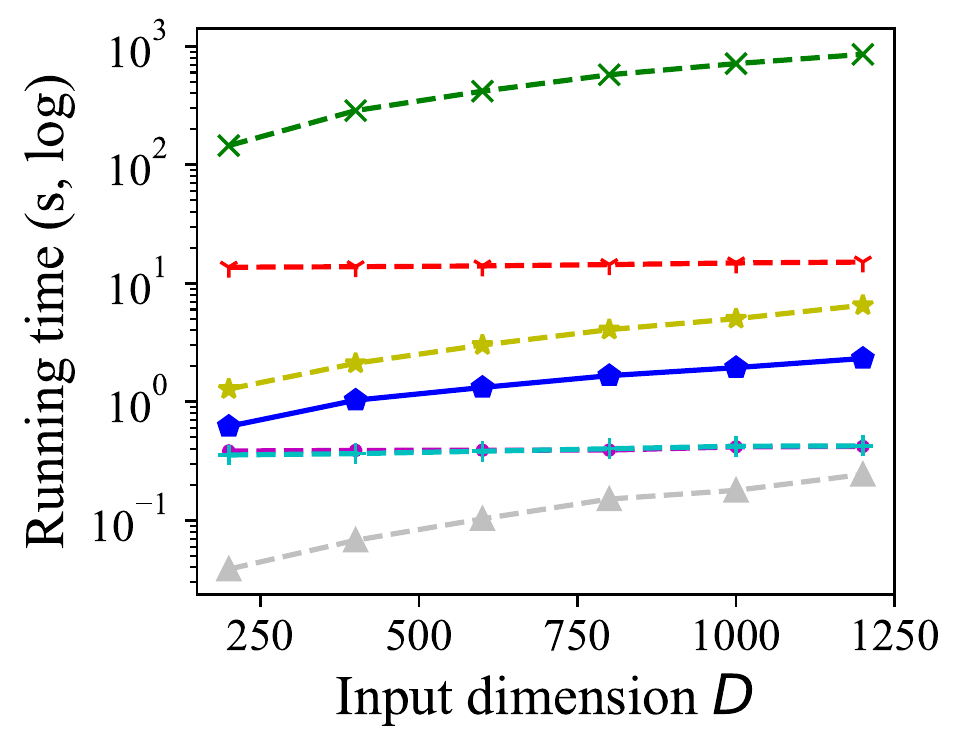}\vspace{-1.75ex}}
\subcaptionbox{\centering Time w.r.t. $T$.
    \label{subfig:time-T}}
{\includegraphics[width=.315\linewidth]{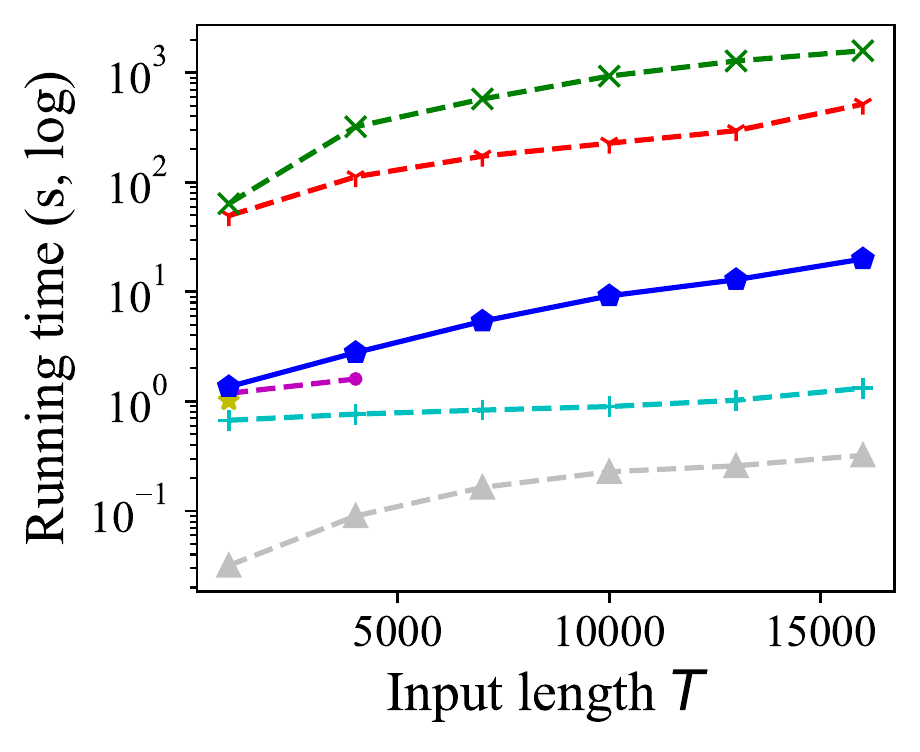}\vspace{-1.75ex}}
\captionsetup[subfigure]{labelformat=empty}
\\ \vspace{-0.5ex}
\subcaptionbox{
    \label{subfig:legend_acc_time}}\centering
{\includegraphics[width=.93\linewidth]{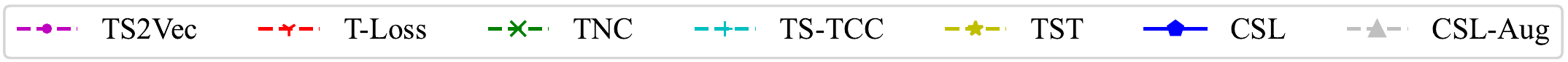}}
\vspace{-6.6ex}
    \caption{\rev{Training time per epoch of varying input size ($N$), dimension ($D$) and series length ($T$) on InsectWingbeat, DuckDuckGeese and EigenWorms respectively.}}
    \label{fig:scalability}
    \vspace{-3.9ex}
\end{figure}

\rev{Although our main goal is to improve the representation quality of URL, we show that the running time of the proposed CSL is also less than or comparable to the URL baselines. We first assess the accuracy with respect to the training time using two medium-sized datasets. As shown in Fig.~\ref{fig:acc-time}, CSL and TS2Vec achieve the same or higher accuracy using much less time, showing that they are faster to train than the other URL methods, while CSL is also faster than TS2Vec. Next, we evaluate the training time per epoch on InsectWingbeat, DuckDuckGeese and EigenWorms, the UEA datasets with the largest input size, dimension and series length. The results are shown in Fig.~\ref{fig:scalability}a-\ref{fig:scalability}c respectively. TS-TCC runs fast in most cases. TS2Vec is also time-efficient, but it cannot scale to large length $T$ due to high memory consumption (Fig.~\ref{subfig:time-T}). TST is slower than CSL for high-dimensional time series (Fig.~\ref{subfig:time-D}) and runs out of memory for large $T$ (Fig.~\ref{subfig:time-T}). T-Loss and TNC, though have smaller time complexity, are much slower than the others with considerably large $N$, $D$ and $T$. The reason is that they consist of many sequential operations which cannot be sped up with GPUs. CSL is fairly efficient among the URL methods in terms of running time per epoch. More importantly, CSL can be faster to train as we have illustrated above as it converges using less number of epochs. Besides, we observe that the time spent on data augmentation (CSL-Aug) is very little during the CSL training.}

\vspace{-0.2ex}
\section{Conclusion}
This paper presents a novel URL framework named CSL, which leverages contrastive learning for MTS-specific representation. Particularly, we design a unified shapelet-based encoder and an objective with multi-grained contrasting and multi-scale alignment to capture information in various time ranges. We also build a data augmentation library including diverse types of methods to improve the generality. Extensive experiments on tens of real-world datasets demonstrate the superiority of CSL over the baselines on downstream classification, clustering, and anomaly detection tasks.


\vspace{-0.2ex}
\begin{acks}
This paper was supported by NSFC grant (62232005, 62202126) and The National Key Research and Development Program of China (2020YFB1006104).
\end{acks}
\balance
\bibliographystyle{ACM-Reference-Format}
\bibliography{references}

\end{document}